\documentclass[pdflatex,sn-mathphys-num]{sn-jnl}

\usepackage{orcidlink}
\usepackage{graphicx}%
\usepackage{multirow}%
\usepackage{amsmath,amssymb,amsfonts}%
\usepackage{amsthm}%
\usepackage{mathrsfs}%
\usepackage[title]{appendix}%
\usepackage{xcolor}%
\usepackage{booktabs}%
\usepackage{anyfontsize} 
\usepackage[range-units=single,per-mode=symbol]{siunitx}
\usepackage{colortbl}
\usepackage{svg}
\usepackage{subcaption}
\usepackage{hyperref}
\usepackage{placeins} 

\newcommand{\shadecell}[2]{%
	\ifnum#1>0
		\cellcolor{green!#2}#2
	\else
		\cellcolor{red!#2}#2
	\fi
}

\usepackage{acro}
\newcommand{\newac}[2]{\DeclareAcronym{#1}{short=#1,long=#2}}
\newac{AI}      {Artificial Intelligence}
\newac{ASTRA}   {Symposium on Advanced Space Technologies in Robotics and Automation}
\newac{CNES}    {French Space Agency}
\newac{CNN}     {Convolutional Neural Network}
\newac{DEM}     {Digital Elevation Model}
\newac{DoF}     {Degrees of Freedom}
\newac{DLSS}    {Deep Learning Super Sampling}
\newac{ESA}     {European Space Agency}
\newac{ESTEC}   {European Space Research and Technology Centre}
\newac{EUSPA}   {European Union Agency for the Space Programme}
\newac{ExoTeR}  {ExoMars Testing Rover}
\newac{FM*}     {Heuristic Fast Marching}
\newac{FMM}     {Fast Marching Method}
\newac{FOG}     {Fibre Optic Gyro}
\newac{FSR}     {FidelityFX Super Resolution}
\newac{FTS}     {Force Torque Sensor}
\newac{GNC}     {Guidance{,} Navigation{,} and Control}
\newac{GNSS}    {Global Navigation Satellite System}
\newac{HDPR}    {Heavy Duty Planetary Rover}
\newac{HRI}     {Human-Robot Interaction Laboratory}
\newac{HR}      {High Resolution}
\newac{IMU}     {Inertial Measurement Unit}
\newac{JFR}     {Journal of Field Robotics}
\newac{LRO}     {Lunar Reconnaissance Orbiter}
\newac{MER}     {Mars Exploration Rover}
\newac{ML}      {Machine Learning}
\newac{MSL}     {Mars Science Laboratory}
\newac{MSR}     {Mars Sample Return}
\newac{MaRTA}   {Martian Rover Testbed for Autonomy}
\newac{NASA}    {National Aeronautics and Space Administration}
\newac{NN}      {Neural Network}
\newac{OBC}     {On Board Computer}
\newac{ORL}     {Orbital Robotics Laboratory}
\newac{OUM}     {Ordered Upwind Method}
\newac{PCA}     {Principal Component Analysis}
\newac{PRL}     {Planetary Robotics Laboratory}
\newac{PTU}     {Pan and Tilt Unit}
\newac{RBF}     {Radial Basis Function}
\newac{RTK}     {Real-Time Kinematics}
\newac{SFR}     {Sample Fetching Rover}
\newac{SLAM}    {Simultaneous Localization and Mapping}
\newac{SPA}     {Sense{,} Plan{,} Act}
\newac{SR}      {Super Resolution}
\newac{SVD}     {Singular Value Decomposition}
\newac{SVM}     {Support Vector Machine}
\newac{SVC}     {Support Vector Classifier}
\newac{TRL}     {Technology Readiness Level}
\newac{UMAP}    {Uniform Manifold Approximation and Projection for Dimension Reduction}
\newac{UMA}     {University of Málaga}
\newac{VO}      {Visual Odometry}
\newac{CoM}     {Center of Mass}
\DeclareAcronym{RND}{short=R\&D,long=Research and Development}
\DeclareAcronym{ROS2}{short=ROS~2,long=Robot Operating System 2}
\DeclareAcronym{TSNE}{short=t-SNE,long=t-distributed Stochastic Neighbor Embedding}

\begin{document}

\title[FTS field assessment]{Field Assessment of Force Torque Sensors for Planetary Rover Navigation}

\newcommand{\mailorcid}[1]{{\small (\orcidlink{#1}~#1)}}

\author*[1]{\fnm{Levin} \sur{Gerdes}\orcidlink{0000-0001-7648-8928}}\email{gerdes@uma.es \mailorcid{0000-0001-7648-8928}}
\author[1]{\fnm{Carlos} \sur{Pérez del Pulgar}\orcidlink{0000-0001-5819-8310}}\email{carlosperez@uma.es \mailorcid{0000-0001-5819-8310}}
\author[1]{\fnm{Raúl} \sur{Castilla Arquillo}\orcidlink{0000-0003-4203-8069}}\email{raulcastar@uma.es \mailorcid{0000-0003-4203-8069}}
\author[2]{\fnm{Martin} \sur{Azkarate}\orcidlink{0000-0003-3284-5422}}\email{martinazkarate@esa.int \mailorcid{0000-0003-3284-5422}}

\affil*[1]{{\small \orgdiv{Space Robotics Lab of the Department of Systems Engineering and Automation}, \orgname{University of Malaga}, \orgaddress{\street{Calle Dr.\ Ortiz Ramos s/n}, \city{Malaga}, \postcode{29001}, \state{Andalusia}, \country{Spain}}}}
\affil[2]{{\small \orgdiv{Planetary Robotics Lab of the Automation and Robotics Section}, \orgname{European Space Agency}, \orgaddress{\street{Keplerlaan 1}, \city{Noordwijk}, \postcode{2201 AZ}, \state{South Holland}, \country{The Netherlands}}}}

\abstract{Proprioceptive sensors on planetary rovers serve for state estimation
	and for understanding terrain and locomotion performance. While inertial
	measurement units (IMUs) are widely used to this effect, force-torque
	sensors are less explored for planetary navigation despite their potential
	to directly measure interaction forces and provide insights into traction
	performance. This paper presents an evaluation of the performance and use
	cases of force-torque sensors based on data collected from a six-wheeled
	rover during tests over varying terrains, speeds, and slopes. We discuss
	challenges, such as sensor signal reliability and terrain response accuracy,
	and identify opportunities regarding the use of these sensors. The data is
	openly accessible and includes force-torque measurements from each of the
	six-wheel assemblies as well as IMU data from within the rover chassis. This
	paper aims to inform the design of future studies and rover upgrades,
	particularly in sensor integration and control algorithms, to improve
	navigation capabilities.}

\keywords{field trial, proprioception, locomotion, terrain analysis}
\pacs[Categories]{(Other) Field assessment, (4), (11)}
\pacs[MSC Classification]{62P30, 68T40, 70Q05}

\maketitle

\section{Introduction}\label{sec:introduction}

Planetary rovers are mobile laboratories, equipped with scientific instruments
to explore remote locations on distant celestial bodies. Navigation is the
critical function that distinguishes them from stationary landers allowing them
to investigate multiple targets across versatile terrains. However, this comes
at the expense of decreased payload mass, a lower energy capacity, and increased
risk due to the traversal itself. In the future, rovers equipped only with
engineering sensors but without their own scientific instruments are a
possibility. The \ac{SFR}, formerly part of the \ac{MSR} campaign, was an
example for such a model. The rover's purpose was to quickly navigate the
Martian surface, locate sample tube caches left behind by the Mars 2020 rover,
and return them to the Mars Ascent vehicle. Rovers traverse unstructured
terrains and can be neither repaired, nor moved or otherwise reset into an
operable state in case of fatal hardware failures. Hence, they rely on
conservative planning by operators on Ground as well as their sensor suite and
on-board capabilities in more autonomous modes.

The sensors on which the operators as well as the rover base their planning can
be categorized as either (1) exteroceptive sensors, such as optical (stereo)
cameras or time-of-flight cameras, or (2) proprioceptive sensors such as
\acp{IMU}, fiber optic gyroscopes, or motor current. \ac{FTS} have already
reached a high \ac{TRL} with example use cases in robotic manipulators, in
terrestrial as well as space applications \cite{ati-perseverance}, and legged
robots. However, they are not commonly found on the locomotion systems of
planetary, wheeled, rovers. This warrants closer investigation as there may be a
signification opportunity for innovation.

\begin{figure}[htb]
	\centering
	\includegraphics[width=0.5\linewidth]{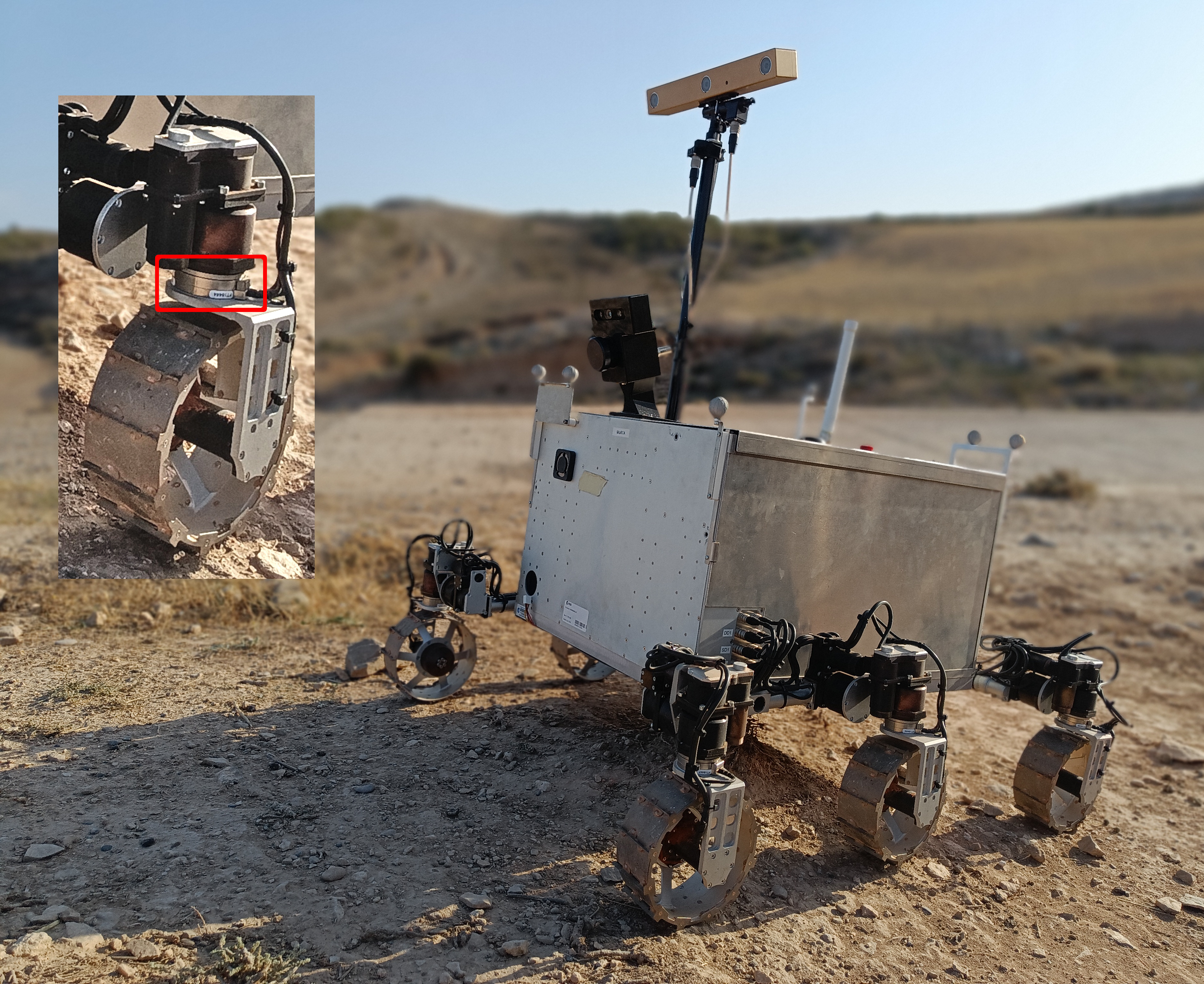}
	\caption{ESA's rover testbed \acs{MaRTA} during the collection of the
		analogue planetary exploration dataset BASEPROD in Bardenas Reales,
		Spain \cite{Baseprod}. The \acp{FTS} can be seen in each of the six legs
		with a close-up highlighting the sensor position relative to the wheel
		axle.}
	\label{fig:marta}
\end{figure}

In this paper, we explore the possibilities offered by the inclusion of
\acp{FTS} in mobile, wheeled robot bases in planetary exploration. Our research
aims to investigate how \acp{FTS} can enhance rover navigation in planetary
scenarios.
In July of 2023, we conducted field trials using \ac{MaRTA}
(\autoref{fig:marta}), one of the European Space Agency's rover testbeds
\cite{MartaExoter}, which is equipped with six ATI mini45 \acp{FTS}, in the
Bardenas Reales semi-desert, an analogue site to Martian and Lunar exploration
\cite{Baseprod}. We evaluate the applications of the \acp{FTS} based on these
experiments, focusing on navigation and traversability-related tasks such as
terrain classification and applicability for drawbar pull estimation with real
data from unstructured terrain.

In \ac{MaRTA}, the \acp{FTS} are mounted above the wheels rather than at the
wheel hubs. This is similar to the mounting position used by the 32 \ac{DoF}
robot RoboSimian \cite{Reid2020a,Reid2020b} which is primarily exploiting the
vertical, gravity-aligned forces. This mounting choice and its trade-offs will
be discussed in detail.

Through our field experiments with the \ac{MaRTA} rover and the analyses in this
paper, we offer an insight into practical challenges and potential applications
of \acp{FTS} in planetary rover systems. Our findings provide guidance and
pointers for researchers in determining the value of incorporating \acp{FTS} in
their mobile robot bases, particularly in unstructured terrains.  The main value
of this paper consists of sharing insights into the configuration of \ac{MaRTA},
with both its benefits and limitations, to inform interested readers which
aspects of this configuration are or are not worth incorporating into their
designs and where we see the need for further refinement.

\section{Related Work}

This section explains work related to the different application areas of
\acp{FTS} within robotics.

\acp{FTS} are widely used in manipulators, both on Earth and in space. On Earth,
\acp{FTS} are used for force-torque control and compliance modes. This enables
safer operation with a decreased risk of damaging components with which the
robot interacts and enables higher tolerances in industrial applications
\cite{kuka-ftcontrol}. The Perseverance rover features an \ac{FTS} on its sample
caching mechanism for collision detection and active force control
\cite{ati-perseverance}. In the ARCHES Analog 1 experiment on Mount Etna in
2022, the team demonstrated the usefulness of \ac{FTS} for operators and
sample-picking tasks in Lunar scenarios with astronauts in the loop
\cite{Panzirsch2022}.

However, the use of \acp{FTS} for navigation purposes in planetary rover
mobility systems is less researched and needs further exploration. The principal
use cases are monitoring whether the wheel is stuck by comparing the motor
current and torque at the wheel axle. In \cite{Toupet2020}, the authors
demonstrate how the experienced torque and correlated motor current can be used
as indications of terrain interaction to extend a rover's mission lifetime.
Later work investigates the use of \acp{FTS} at the axle for terrain
classification \cite{Ugenti2022}.

Typically \acp{IMU} are used for terrain classification
\cite{Brooks2005,Coyle2010,Bai2019,Vulpi2021}. Note how in \cite{Brooks2005},
the authors place the \ac{IMU} in the link connected to the wheel hub to detect
specific vibrations related to the actuators. Other sensors used for
classification include sound \cite{Xue2022} and motor current \cite{Ugenti2022}.

In legged robots, research suggests that \acp{FTS} are more suitable for terrain
classification than vibration-based approaches using \ac{IMU}
\cite{Walas2016,Kolvenbach2019,Bednarek2019}. This makes sense, of course, given
that legs make contact at discrete times instead of staying in contact, like a
wheel, and rolling over the terrain.

To the best of the authors' knowledge, there is no research investigating the
exclusive exploitation of \acp{FTS} for terrain classification for wheeled
rovers. Instead, they appear to be at most used in conjunction with the more
traditional \acp{IMU}. In their work on terrain classification, Ugenti et al.\
compare the performance of a \ac{SVM} and a \ac{NN} with task-specific feature
selection \cite{Ugenti2022}. The work includes both \ac{IMU} and \ac{FTS} but
does not contrast their individual performance.

Another challenge we face when considering approaches for terrain classification
in planetary exploration is that the classes are less distinct than in the bulk
of the research body, which aims to distinguish between grass, pavement, and
concrete. Additionally, the rover should ideally be able to classify terrains in
more challenging applications, while varying speed and driving along a turn or
sideways along a slope for example. The closest we found is a testbed developed
by Yang et al.\ \cite{Yang2015}, which uses an \acs{FTS} at the hub of a car as
well as external, manually operated instruments to record relevant data for
wheel-soil interaction. The authors deplore the lack of similar, realistic
recordings of relevant data in soft terrains.

Note that for our us  case, we also need hard terrains (such as bedrock or
compressed sand) and are not as interested in high speeds and dynamics, as
planetary rovers usually operate in the order of centimeters per second.

Monitoring wheel slip is vital for wheeled vehicles, especially in off-road
scenarios, as cars can dig themselves in. This problem can also occur at the
much lower speeds of planetary rovers. The most prominent example is the loss of
the Spirit rover, which, after more than six years traversing the Martian
surface with a top speed of \SI{3.75}{\cm\per\second}, got trapped in a patch of
sand in 2009 \cite{spirit}. The importance of slip for rovers is also
demonstrated in \cite{Yoshida2002}.

In planetary robotics, slip can be intuitively determined by comparing wheel
speed to vehicle speed. The problem with this approach, however, is that the
vehicle speed must first be determined. Planetary rovers equipped with cameras
would typically estimate the vehicle's speed via visual odometry, such as NASA's
Mars rovers \cite{VisodomJpl} and the upcoming ExoMars rover \cite{Visloc}.
Alternative locomotion modes, such as wheel walking \cite{WheelWalkingAstra,
TimWheelWalking}, have to be treated differently, as the wheel rotation does not
suffice as an input to the slip estimation \cite{Salva2022}.

Drawbar pull indicates how much force is available for a vehicle to pull. This
depends on the vehicle-soil interaction and is related to the slip ratio and
rolling resistance. Higher slip generally leads to lower drawbar pull.
Conversely, lower slip improves the drawbar pull, enhancing the rover's
capability to navigate and maneuver. The edge case of no slip is the exception
where the rover does not move.

Understanding and accurately estimating wheel slip and drawbar pull is therefore
critical, as it directly affects the rover's overall mobility and effective net
traverse in challenging or risky terrains \cite{Patel2010}. In their work on
finding simplified formulas for wheel-soil interaction, Shibly et al.\ used an
\ac{FTS} at the wheel hub of a single-wheel testbed to measure the drawbar pull
of a rigid wheel on deformable soil \cite{SHIBLY20051}. In this work, the soil
is not prepared to ideal conditions and we provide insight into our real-world
recordings. We highlight the challenges brought about by sensor data of a fully
mobile rover traversing unknown terrain at varying speeds and orientations.

\FloatBarrier
\section{Terrain classification}
\label{sec:terrai-classification}

\begin{figure}[htb]
	\centering
	\begin{subfigure}[h]{0.49\linewidth}
		\includegraphics[width=\linewidth]{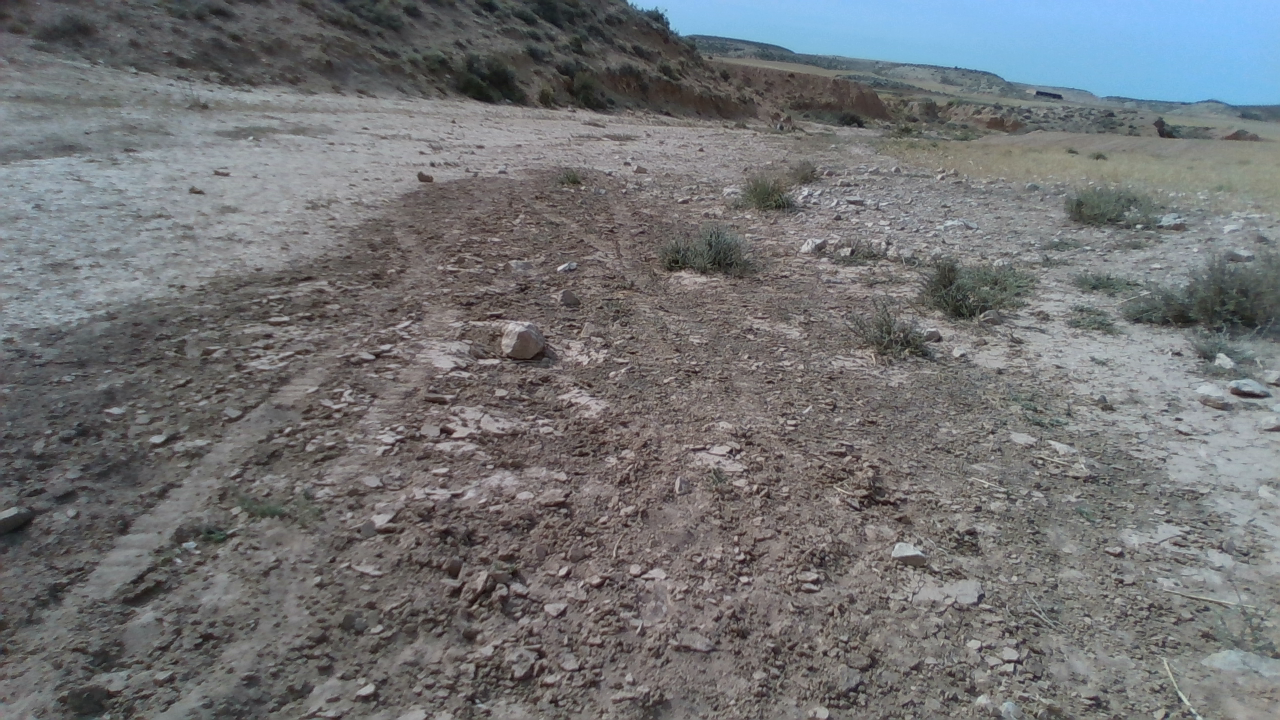}
		\caption{Loose soil in traverse `2023-07-21 17-34-18'.}
		\label{fig:terrain-loose}
	\end{subfigure}
	\begin{subfigure}[h]{0.49\linewidth}
		\includegraphics[width=\linewidth]{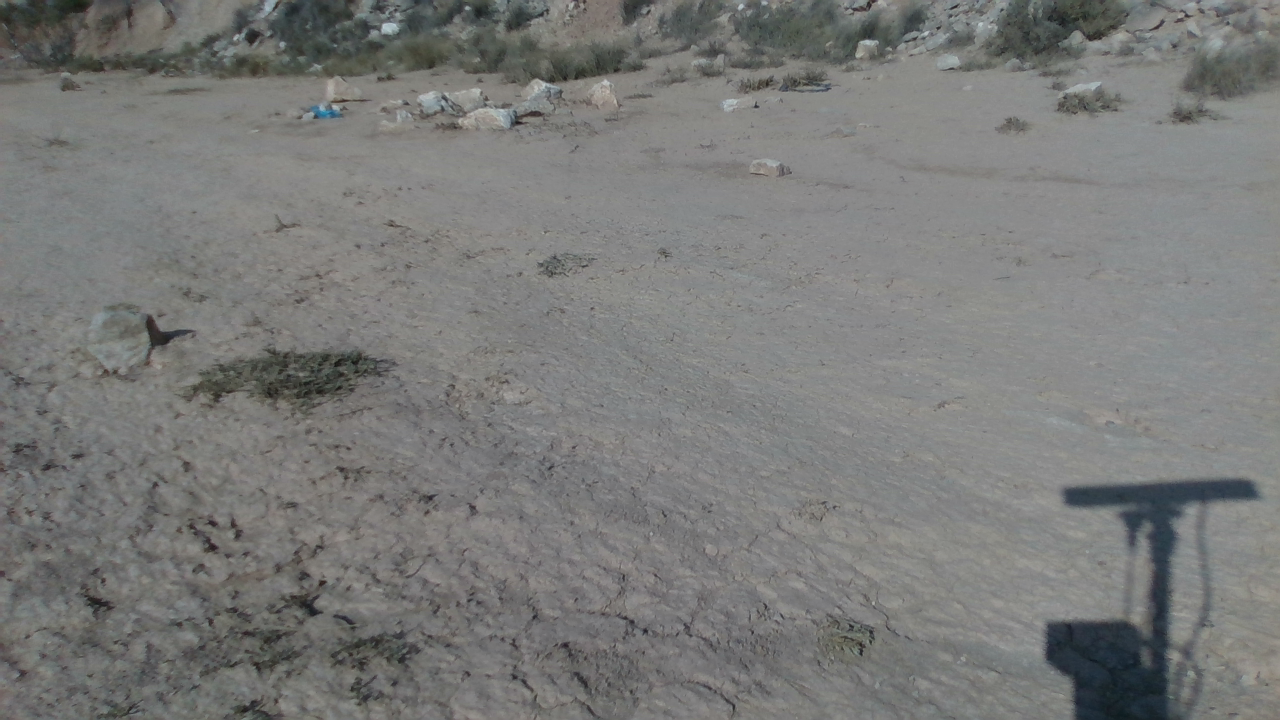}
		\caption{Compressed sand in traverse `2023-07-20 18-12-05'.}
		\label{fig:terrain-compressed}
	\end{subfigure}

	\begin{subfigure}[h]{0.49\linewidth}
		\includegraphics[width=\linewidth]{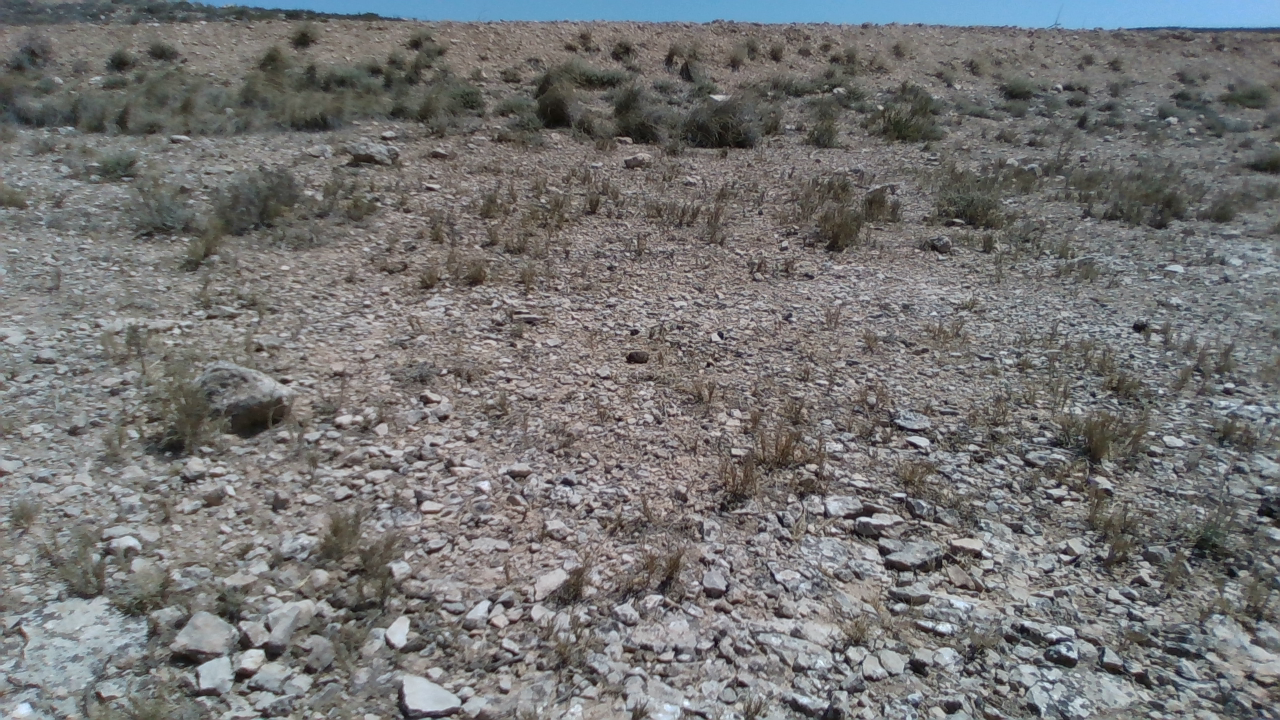}
		\caption{Pebbles in traverse `2023-07-21 12-58-11'.}
		\label{fig:terrain-pebbles}
	\end{subfigure}
	\begin{subfigure}[h]{0.49\linewidth}
		\includegraphics[width=\linewidth]{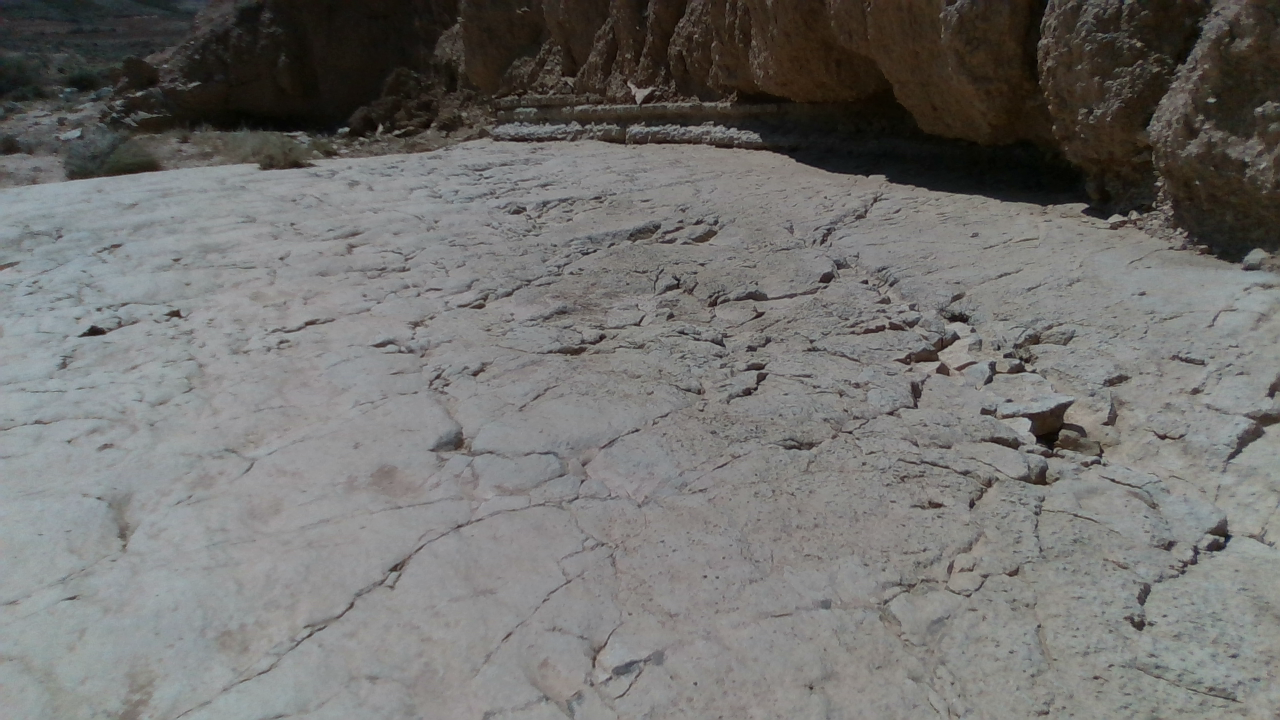}
		\caption{Rock in traverse `2023-07-21 12-58-11'.}
		\label{fig:terrain-rock}
	\end{subfigure}
	\caption{Rover view of the example terrains classified as
		(\subref{fig:terrain-loose})~``loose'',
		(\subref{fig:terrain-compressed})~``compressed'',
		(\subref{fig:terrain-pebbles})~``pebbles'', and
		(\subref{fig:terrain-rock})~``rock''.
		The images were taken with the rover's front-facing RGB-D camera.}
	\label{fig:terrains}
\end{figure}

In the literature, we saw that proprioceptive terrain classification was mainly
tested for straight traverses over flat terrains and constant speeds. In our
test data, which contained slopes, varying speeds (albeit within only
\SIrange{1}{5}{\cm\per\second}), and turns, even point turns, we could still
classify the terrains between four major categories. \autoref{fig:terrains}
shows examples of the categories we considered.

\begin{figure}[htb]
	\centering
	\includegraphics[width=.5\linewidth]{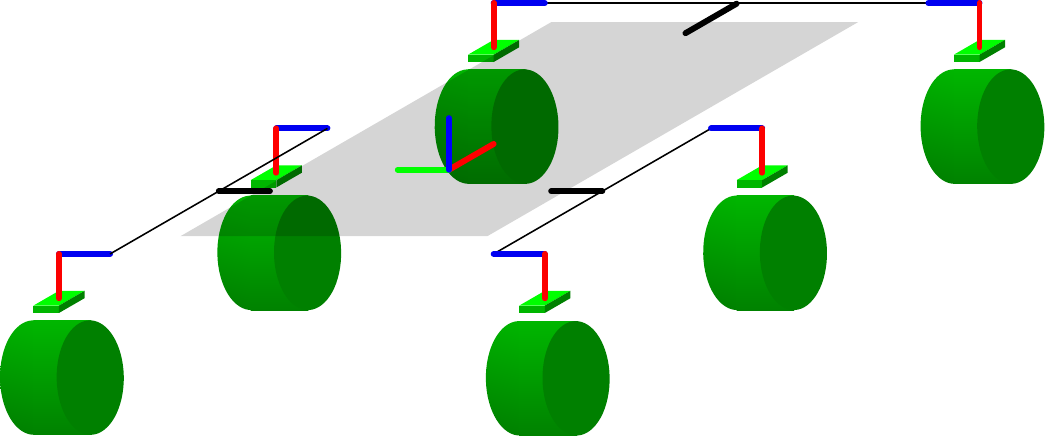}
	\caption{Simplified kinematics of MaRTA's locomotion platform. In this figure,
		MaRTA is headed toward the viewer. The black lines indicate the three
		passive bogies with the thick black lines representing the passive bogie
		joints. The deployment actuators are indicated in blue, steering in red.
		The green boxes represent the \acp{FTS} and the cylinders the wheels.
		The \ac{IMU} and its axes can be seen in the rover's center.}

	\label{fig:kinematics-abstract}
\end{figure}

\acs{MaRTA}'s main components for this work can be seen in the simplified
overview in \autoref{fig:kinematics-abstract}. Unlike single-wheel testbed
setups, for example, where the \acp{FTS} are commonly mounted at the wheel hub,
\acs{MaRTA}'s \acp{FTS} are positioned above just above the wheel bracket,
between the wheel and the steering joint. The figure shows the triple bogie
setup with one \ac{FTS} per leg assembly and that the \ac{IMU} is located on the
base plate of the rover chassis with $x$ pointing backward.

\begin{figure}[htb]
	\includegraphics[width=\linewidth]{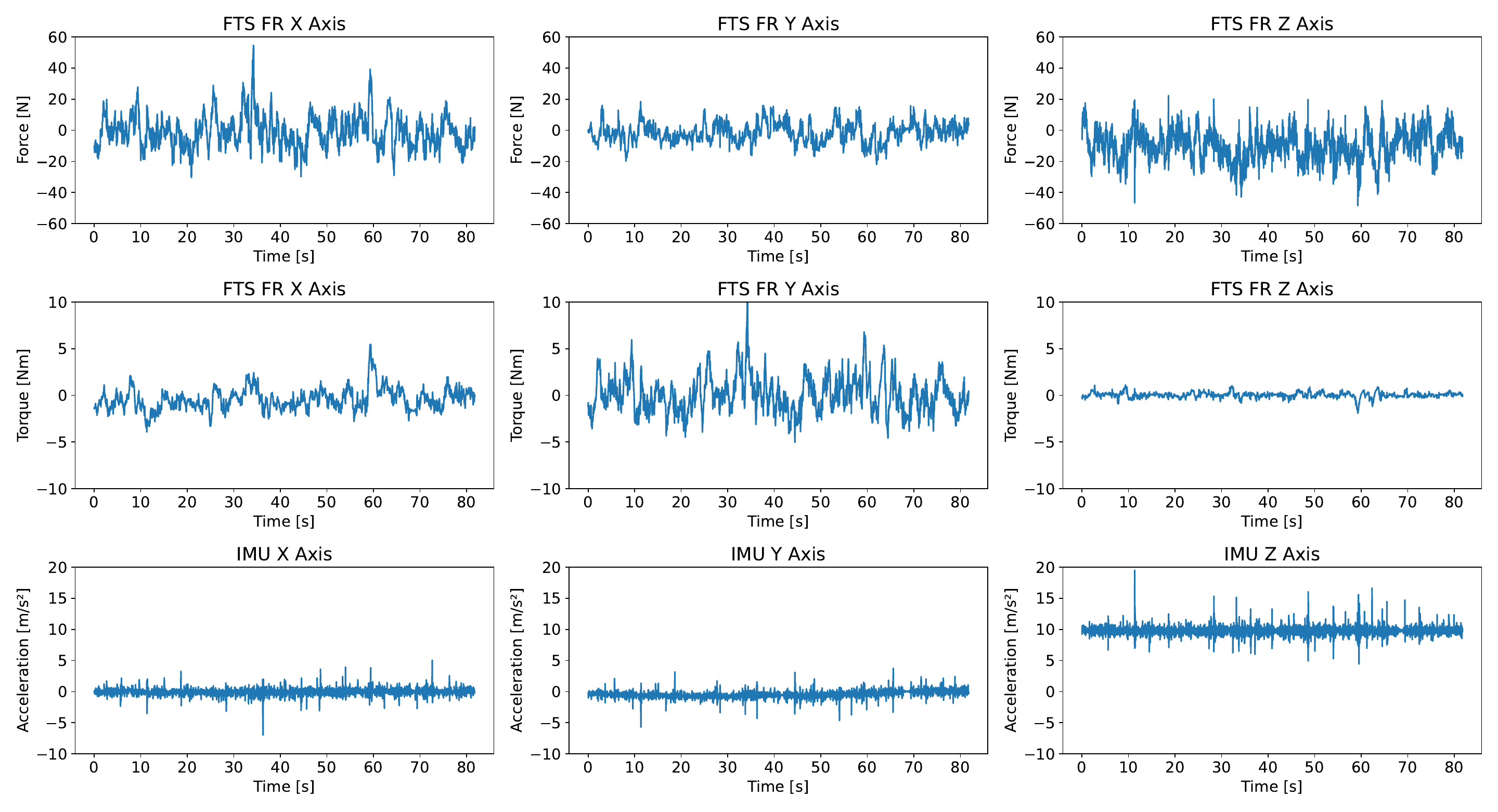}

	\caption{\ac{FTS} and \ac{IMU} data on pebbles. The first row shows front
		right \acs{FTS}'s forces along its three axes, the second row shows the
		same \acs{FTS}'s torques, and the third row visualizes the \ac{IMU}'s
		acceleration data. The three columns correspond to the sensors' $x$,
		$y$, and $z$ axes respectively.}

	\label{fig:fts-data-pebbles}
\end{figure}

\begin{figure}[htb]
	\includegraphics[width=\linewidth]{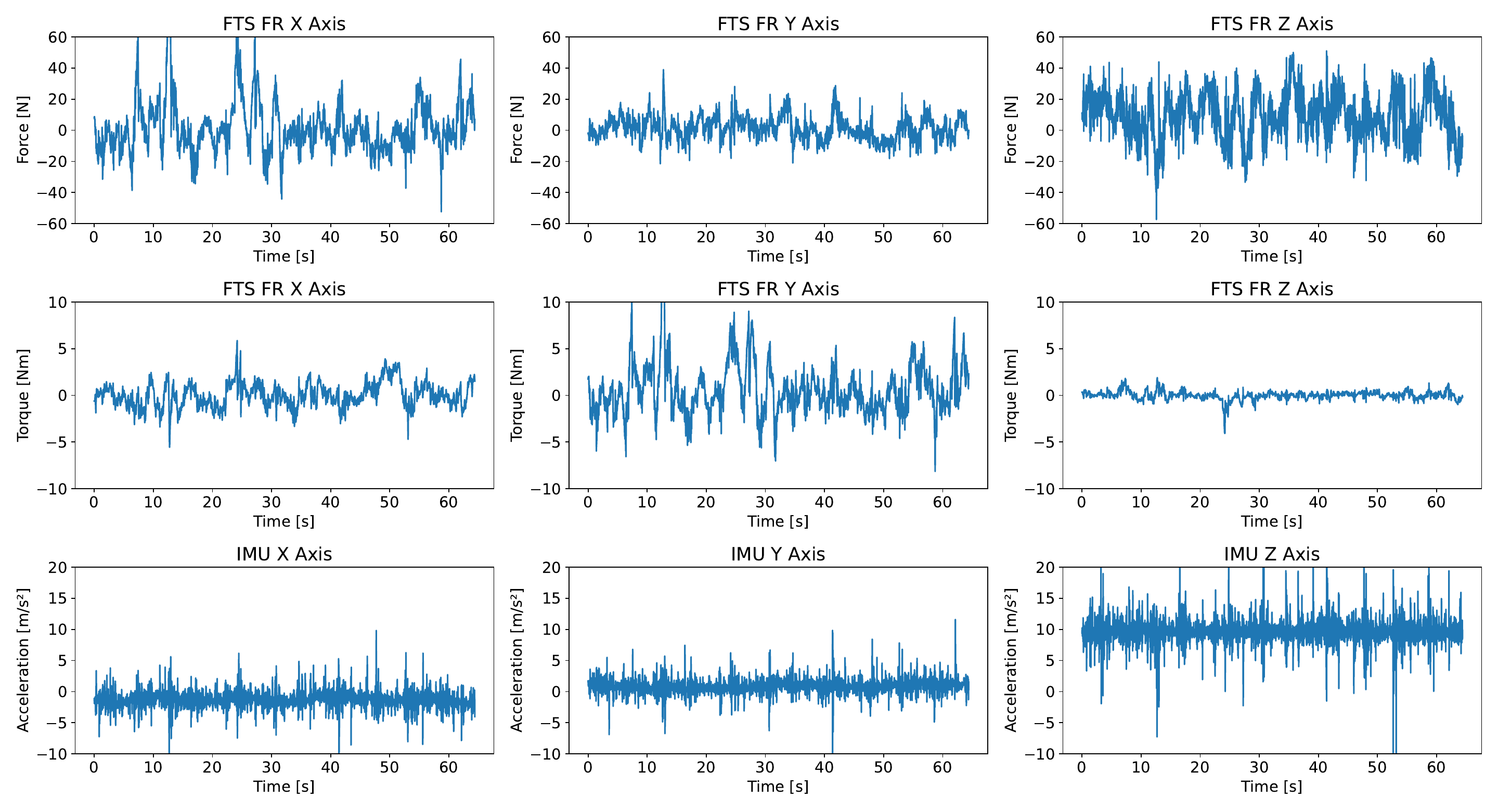}

	\caption{\ac{FTS} and \ac{IMU} data on pebbles. The first row shows front
		right \acs{FTS}'s forces along its three axes, the second row shows the
		same \acs{FTS}'s torques, and the third row visualizes the \ac{IMU}'s
		acceleration data. The three columns correspond to the sensors' $x$,
		$y$, and $z$ axes respectively.}

	\label{fig:fts-data-rock}
\end{figure}

To understand the available signals better, refer to
\autoref{fig:fts-data-pebbles} compared to \autoref{fig:fts-data-rock}. The
figures show plots for the traversal of a dried-out riverbed with pebbles
compared to the traversal of rocky outcrops. The magnitude variation in all
force, torque, and \ac{IMU} acceleration readings are visibly larger during the
outcrop traversal compared to the riverbed. Notice that this effect is
observable although the data is from actual traverses at varying rover
orientations and speeds. At first glance, the difference seems to be less
pronounced in the \ac{FTS} data than in the \ac{IMU} recording.

We identified four terrains that we could visually identify and that appear in
multiple locations and traverses throughout the dataset. To get a better idea of
whether we can use our proprioceptive data for the classification, we use a
subset of the data where these terrains are distinct and where we have multiple
meters within the same type of terrain. Notice, however, that we do not have
perfectly level rover orientations (in some parts we are even climbing slopes,
sometimes traversing them diagonally), and the rover speed is not constant
either. We are interested in the classification of data that is as
representative of real rover traverses as possible while still being able to
verify which terrain is which. To the best of the authors' knowledge, similar
work is primarily done on short, straight traverses, and, if possible, on flat
terrains. Additionally, the `pebbles' class is arguably overlapping with the
`compressed sand' and the `rock' class.

\begin{figure}[htb]
	\centering
	\includegraphics[width=0.5\linewidth]{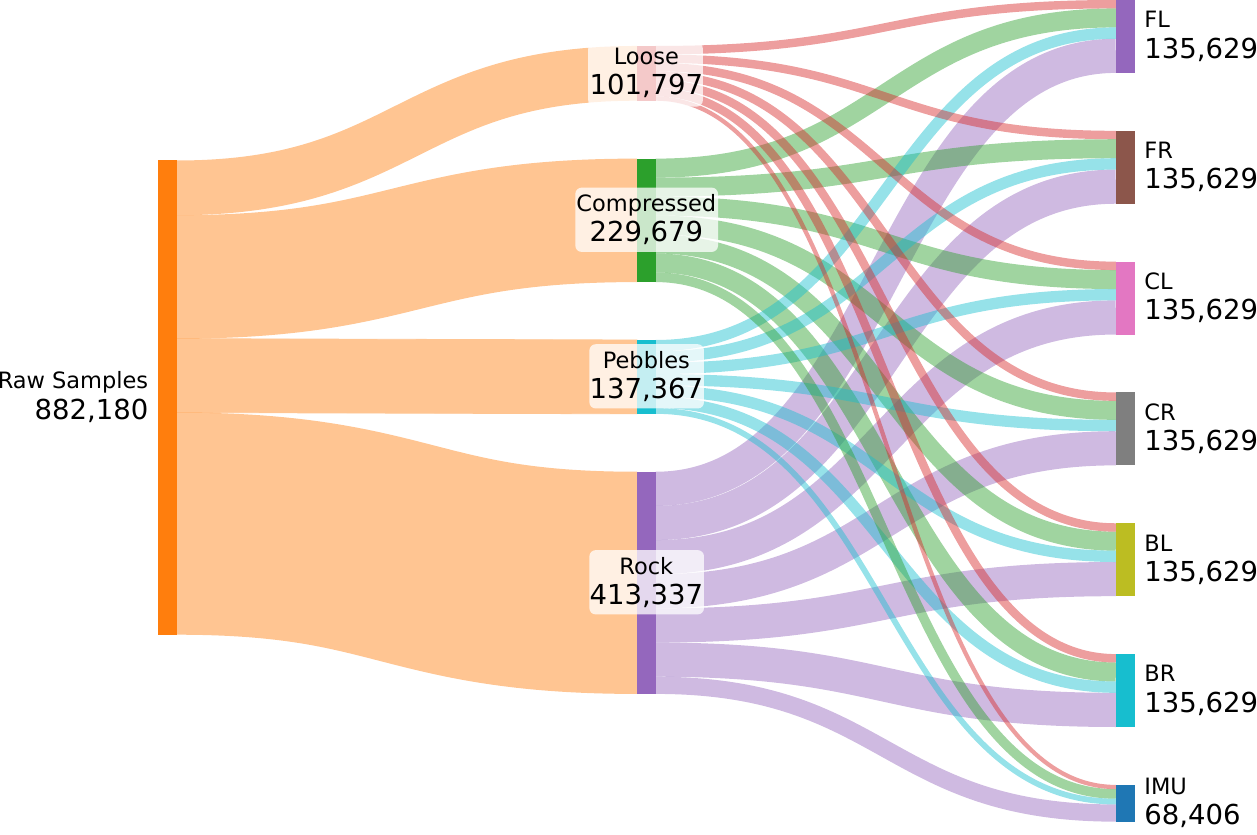}
	\caption{Breakdown of the labeled sample data over terrains and sensors,
		visualizing the number of data points before creating the statistics
		over 1-second-long sliding windows. The graph was created using
		SankeyMATIC \cite{sankeymatic}.}
	\label{fig:sankey}
\end{figure}

\autoref{fig:sankey} visualizes the distribution of the considered data points
per terrain and sensor. We would like to classify the sensor data for each time
instance, e.g., for each second of a traverse. Since the sensors were not
synchronized, we chose to use a 1-second-long sliding window during which we
considered the data from all sensors. After our first inspection of
\autoref{fig:fts-data-pebbles} and \autoref{fig:fts-data-rock}, we chose to not
only use each sensor's average value, but also other measures as meaningful
input for any classification. As such, a processed sample contains each sensor's
mean, median, minimum, maximum, and standard deviation per second.
\autoref{tab:classification-sample-overview} shows how many data points we used
and how many samples this yields for each of the terrain classes.

\begin{table}[htb]
	\centering
	\begin{tabular*}{\linewidth}{@{\extracolsep\fill}lcccccc}\toprule
		&& \multicolumn{5}{c}{Terrain} \\\cmidrule{3-7}
		\multicolumn{2}{c}{Sample type}   & Loose     & Compressed & Pebbles   & Rock      & Total     \\\midrule
		\multirow{2}{*}{Raw} & Count & 101,797 & 229,679 & 137,367 & 413,337 & 882,180 \\\cmidrule{2-2}
		& Share & \SI{11.54}{\%} & \SI{26.04}{\%} & \SI{15.57}{\%} & \SI{46.85}{\%} & \SI{100.00}{\%} \\\cmidrule{1-2}
		\multirow{2}{*}{Processed} & Count & 157     & 354     & 212     & 637     & 1,360   \\\cmidrule{2-2}
		& Share & \SI{11.54}{\%} & \SI{26.03}{\%} & \SI{15.59}{\%} & \SI{46.84}{\%} & \SI{100.00}{\%}     \\\bottomrule
	\end{tabular*}
	\caption{Overview of the amount of data used for classification in total and
		per class. A `raw' data point is a single data row from one of the
		sensors, e.g., one set of \ac{IMU} accelerations and orientations with
		corresponding timestamp. The processed sample is the set of all sensors'
		statistics (mean, min, etc.) over one second.}
	\label{tab:classification-sample-overview}
\end{table}

\begin{figure}[htb]
	\centering

	\begin{subfigure}[h]{0.49\linewidth}
		\includegraphics[width=\linewidth]{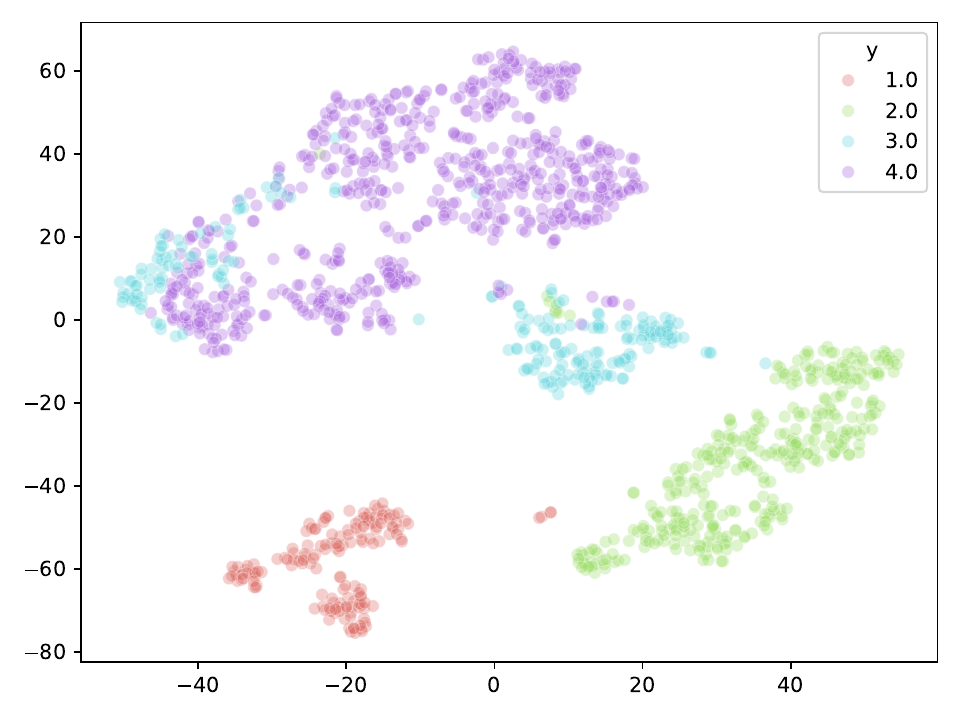}
		\label{fig:clustering-imu}
		\caption{\ac{TSNE} result for \ac{IMU} data.}
	\end{subfigure}
	\begin{subfigure}[h]{0.49\linewidth}
		\includegraphics[width=\linewidth]{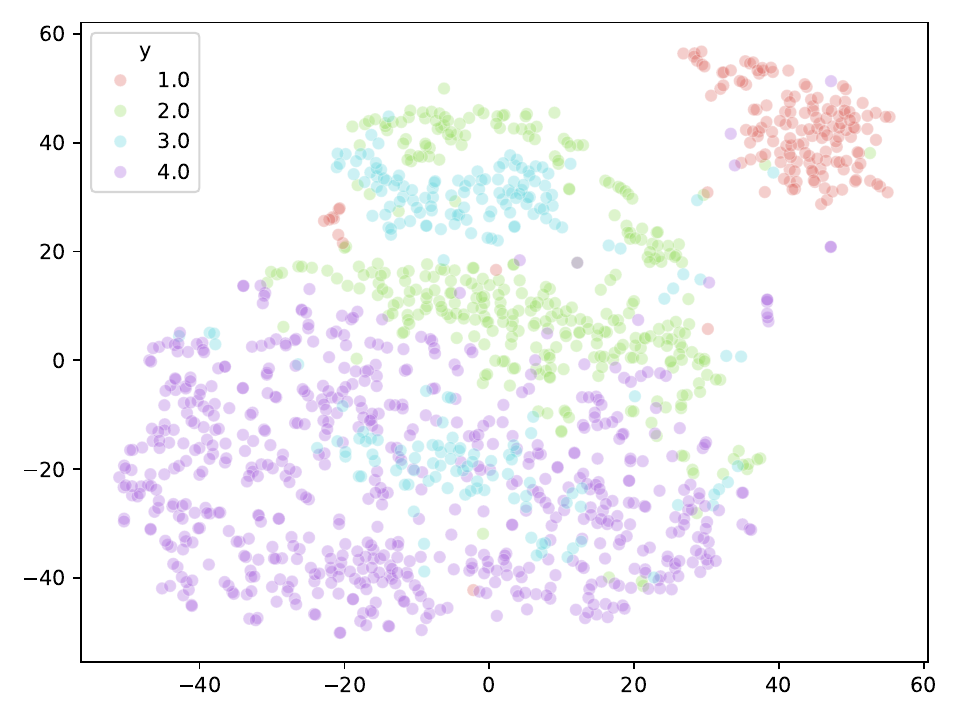}
		\label{fig:clustering-fts}
		\caption{\ac{TSNE} result for \ac{FTS} data.}
	\end{subfigure}

	\caption{Low-dimensional representation and clustering of the \ac{IMU} and
		\ac{FTS} statistics created using \ac{TSNE}. Clusters 1 through 4 are
		terrain classes loose soil, compressed sand, pebbles, and rock
		respectively.}
	\label{fig:tsne}

\end{figure}

We analyzed the data using \ac{PCA}, Truncated \ac{SVD}, and \ac{TSNE} before
exploring the application of \acp{SVM} or \acp{NN}.  We considered a range of 15
to 225 input parameters, computed for 1-second-long samples of known terrains
(with slightly varying speeds). The 15 parameters are already reached when using
only the \ac{IMU}'s acceleration along its three axes (times five statistical
values thereof). The 225 parameters are reached when using all sensors: IMU
accelerations and the six \acsp{FTS}' forces and torques in three axes as well
as a derived value of the \acsp{FTS}' force in $x$ over its torque in $z$.
Please refer to the later section on drawbar pull why we consider this useful.
\autoref{fig:tsne} shows the \ac{TSNE} visualization of the \ac{IMU} and
\ac{FTS} data.

We gather from these graphs that terrain classification should be possible with
\ac{FTS}, even though \ac{IMU} seems better suited for the types of terrains we
are interested in. Note that we use these plots only as a first step to
visualize the data, we are not drawing direct conclusions from them, because, as
shown by Chari et al.\ \cite{Chari2023}, visualizations of high-dimensional data
in only two or three dimensions might lead to wrong conclusions and should not
be mistaken as or used for clustering. The researchers illustrate that neither
relative distance nor density are preserved (shown for both \ac{TSNE} and
\ac{UMAP} \cite{umap}) and how sensitive the clustering results are wrt.\ the
number of neighbors, ``perplexity'' in the case of \ac{TSNE}. Additionally,
\ac{TSNE} does not yield an explicit function from input space to the mapped
space \cite{tsne-faq}. We use the visualization only to get and show a first
impression of the data to estimate feasibility, but the mapping from the input
space to the classes has to be achieved in a different way.

\subsection{Classification with Support Vector Machines (SVMs)}

The first method we used for classification is \acfp{SVM}. An \ac{SVM}, or in
this instance more precisely a \ac{SVC}, finds a set of hyperplanes to best
separate instances of different classes. Since \ac{SVM} were originally
conceived for binary classification, we use Scikit-Learn's \ac{SVC}
implementation\footnote{\url{https://scikit-learn.org/1.5/modules/svm.html\#multi-class-classification}}
which enables multi-class classification via either a ``one-vs-one'' or a
``one-vs-rest'' approach \cite{scikit-learn}. This means that an \ac{SVC} for an
$N$-class classification trains either $N \cdot \frac{N-1}{2}$ many classifiers,
each to distinguish between two classes, or $N$ classifiers, each comparing one
class to all others.

In this section, we present a comparison of the prediction accuracies when using
only \ac{IMU}, only \ac{FTS}, and when using both \ac{IMU} and \ac{FTS} data.
For each of these cases, we used Scikit-Learn \cite{scikit-learn} to setup a
grid search over the most common \ac{SVC} kernels and parameters:

\begin{itemize}
	\item $C \in \{0.1, 1, 10, 100\}$
	\item $\gamma \in \{1, 0.1, 0.01, 0.001\}$
	\item kernel $\in$ \{linear, \ac{RBF}, polynomial, sigmoid\}
\end{itemize}

The kernel is the principal factor for how the classification boundaries are
determined with the linear kernel being the least expressive while the other
kernels can create more complex hyperplanes. $C$ is a regularization parameter
which is often set as a negative power of 10. The smaller $C$, the stronger the
regularization. $\gamma$ can be found in the non-linear kernels' definitions and
thus has an effect depending on the kernel. For an intuition behind the effect
of these parameters, we recommend Scikit's visualizations regarding
$C$\footnote{\url{https://scikit-learn.org/1.5/auto_examples/svm/plot_svm_scale_c.html}}
and
kernel\footnote{\url{https://scikit-learn.org/1.5/auto_examples/svm/plot_svm_kernels.html}}.

We experimented with both ``one-vs-one'' and ``one-vs-all'' classification, but
achieved the same accuracies. Hence, we are only reporting the ``one-vs-all''
results below and did not repeat the timing for ``one-vs-one''.

\autoref{tab:svm-res-imu} shows the confusion matrix for an \ac{SVM} classifying
the \ac{IMU} data with the best set of parameters, $C=0.1$, $\gamma=0.1$, and,
notably, a polynomial kernel. We can see that the \ac{SVM} has the highest
accuracy (\SI{96.23}{\percent}) when identifying `rock' and the second highest
accuracy for `loose soil' (\SI{90.91}{\percent}). Intuitively, these are the
furthest apart in terms of vibrations experienced by the rover during a
traverse. The \ac{SVM} mainly struggles in identifying `pebbles'
(\SI{58.21}{\percent}) with \SI{22.39}{\percent} and \SI{17.91}{\percent}
misclassified as `compressed sand' and `rock' respectively.

\begin{table}[htb]
	\begin{tabular*}{\linewidth}{@{\extracolsep\fill}lrrrr}%
		\toprule
		&\multicolumn{4}{c}{Predicted}\\\cmidrule{2-5}
		Actual & Loose & Compressed & Pebbles & Rock\\
		\midrule
		Loose      & \shadecell{1}{90.91} & \shadecell{0}{ 9.09} & \shadecell{0}{ 0.00} & \shadecell{0}{ 0.00} \\
		Compressed & \shadecell{0}{ 0.00} & \shadecell{1}{86.25} & \shadecell{0}{13.75} & \shadecell{0}{ 0.00} \\
		Pebbles    & \shadecell{0}{ 1.49} & \shadecell{0}{22.39} & \shadecell{1}{58.21} & \shadecell{0}{17.91} \\
		Rock       & \shadecell{0}{ 0.63} & \shadecell{0}{ 1.26} & \shadecell{0}{ 1.89} & \shadecell{1}{96.23} \\
		\bottomrule
	\end{tabular*}
	\caption{Row-normalized confusion matrix and accuracies for an \ac{SVM} with
		polynomial kernel on \ac{IMU} data only, with a \SI{25}{\percent} test
		split. Training accuracy was \SI{89.51}{\percent} and test accuracy
		\SI{85.84}{\percent}.}
	\label{tab:svm-res-imu}
\end{table}

When looking into the results for \ac{FTS} data (\autoref{tab:svm-res-fts}), we
saw that a linear kernel yielded the best result (with $C=0.1$ and $\gamma=1$),
with an overall test accuracy improvement from \SI{85.84}{\percent} to
\SI{95.58}{\percent}. The largest improvement can be seen in the accuracy for
the pebble class: only \SI{1.49}{\percent} get misclassified as `compressed' and
the test accuracy for just the pebble class climbs to \SI{85.07}{\percent}.

\begin{table}[htb]
	\begin{tabular*}{\linewidth}{@{\extracolsep\fill}lrrrr}
		\toprule
		&\multicolumn{4}{c}{Predicted}\\\cmidrule{2-5}
		Actual & Loose & Compressed & Pebbles & Rock\\
		\midrule
		Loose      & \shadecell{1}{100.00} & \shadecell{0}{ 0.00} & \shadecell{0}{ 0.00} & \shadecell{0}{ 0.00} \\
		Compressed & \shadecell{0}{  1.25} & \shadecell{1}{97.50} & \shadecell{0}{ 0.00} & \shadecell{0}{ 1.25} \\
		Pebbles    & \shadecell{0}{  0.00} & \shadecell{0}{ 1.49} & \shadecell{1}{85.07} & \shadecell{0}{13.43} \\
		Rock       & \shadecell{0}{  0.63} & \shadecell{0}{ 0.00} & \shadecell{0}{ 1.26} & \shadecell{1}{98.11} \\
		\bottomrule
	\end{tabular*}
	\caption{Row-normalized confusion matrix and accuracies for an \ac{SVM} with
        linear kernel on \ac{FTS} data only, with a \SI{25}{\percent} test
        split. Training accuracy was \SI{100}{\percent} and test accuracy
        \SI{95.58}{\percent}.}
	\label{tab:svm-res-fts}
\end{table}

Including the \ac{IMU} in combination with the \ac{FTS} inputs, we see results
comparable to only using the \ac{FTS} (see \autoref{tab:svm-res-all}). First,
the grid search yielded the same parameters: $C=0.1$, $\gamma=1$, and a linear
kernel. The accuracies for the `loose', `compressed', and `rock' classes are
unchanged, suggesting roughly the same rules were learned here. Only in the
`pebbles' class do we notice an improvement: fewer `pebbles' samples get
misclassified as `rock' (\SI{11.94}{\percent} versus \SI{13.43}{\percent}).

\begin{table}[htb]
	\begin{tabular*}{\linewidth}{@{\extracolsep\fill}lrrrr}%
		\toprule
		&\multicolumn{4}{c}{Predicted}\\\cmidrule{2-5}
		Actual & Loose & Compressed & Pebbles & Rock\\
		\midrule
		Loose      & \shadecell{1}{100.00} & \shadecell{0}{ 0.00} & \shadecell{0}{ 0.00} & \shadecell{0}{ 0.00} \\
		Compressed & \shadecell{0}{  1.25} & \shadecell{1}{97.50} & \shadecell{0}{ 0.00} & \shadecell{0}{ 1.25} \\
		Pebbles    & \shadecell{0}{  0.00} & \shadecell{0}{ 1.49} & \shadecell{1}{86.57} & \shadecell{0}{11.94} \\
		Rock       & \shadecell{0}{  0.63} & \shadecell{0}{ 0.00} & \shadecell{0}{ 1.26} & \shadecell{1}{98.11} \\
		\bottomrule
	\end{tabular*}
	\caption{Row-normalized confusion matrix and accuracies for an \ac{SVM} with
		linear kernel on \ac{IMU} and \ac{FTS} data with a \SI{25}{\percent}
		test split. Training accuracy was \SI{100}{\percent} and test accuracy
		\SI{95.87}{\percent}.}
	\label{tab:svm-res-all}
\end{table}

At first, it is surprising to see that the `pebbles' accuracy improves by
including the \ac{IMU}, because we have just seen that the classifier using only
\ac{IMU} struggles the most with this class. Still, we provide the classifier
with additional data samples by including more data sources.

\subsection{Classification with Neural Networks (NNs)}

\begin{figure}[htb]
	\centering
	\includegraphics[width=\linewidth]{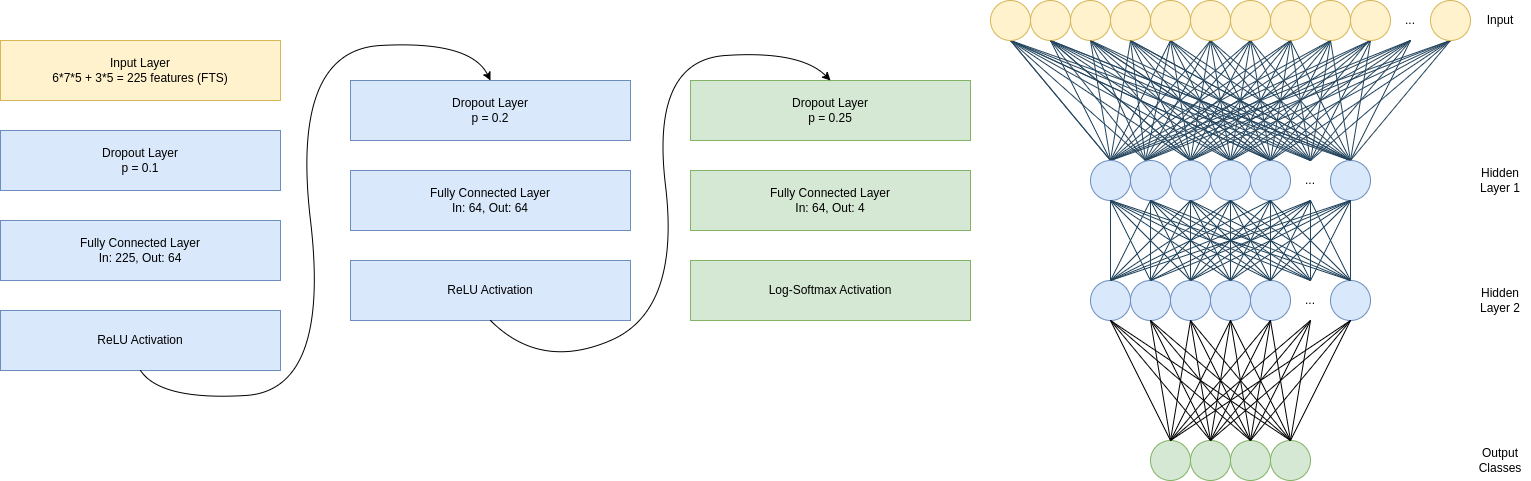}
	\caption{Architecture of the neural network used for testing terrain
		classification. The two hidden layers are comprised of 64 units each and
		the number of inputs depends on the basis for classification.}
	\label{fig:nn-arch}
\end{figure}

Using the same 1361 samples from above, we trained a simple neural network with
one hidden layer of 64 units, as shown in \autoref{fig:nn-arch}. The network was
implemented in PyTorch \cite{pytorch} and trained on \SI{85}{\percent} of the
data and tested on the remaining \SI{25}{\percent}. Dropout regularization was
implement with a \SI{10}{\percent} rate for the inputs and \SI{20}{\percent}
between the hidden layers. The chosen minibatch size was 32 and the training ran
for 50 epochs. Similar to the \ac{SVM}, we trained the \ac{NN} on three input
variants: (1) only \ac{IMU}, (2) only \ac{FTS}, and (3) both \ac{IMU} and
\ac{FTS}.

\autoref{tab:nn-res-imu} shows the confusion matrix for training the \ac{NN} on
only the \ac{IMU} data. We can see that, similar to the \ac{SVM}, the \ac{NN}
achieves high accuracies for the `rock' and `loose' classes, both more than
\SI{96}{\percent}, this is even higher score than the \ac{SVM}. But the overall
accuracy of \SI{79.35}{\percent} is less than the \SI{85.84}{\percent} of the
\ac{SVM} for the same data. The misclassification of `pebbles' as `rock' is
greater than \SI{67}{\percent}.

\begin{table}[htb]
	\begin{tabular*}{\linewidth}{@{\extracolsep\fill}lrrrr}%
		\toprule
		&\multicolumn{4}{c}{Predicted}\\\cmidrule{2-5}
		Actual & Loose & Compressed & Pebbles & Rock\\
		\midrule
		Loose      & \shadecell{1}{96.97} & \shadecell{0}{ 3.03} & \shadecell{0}{ 0.00} & \shadecell{0}{ 0.00} \\
		Compressed & \shadecell{0}{ 0.00} & \shadecell{1}{80.00} & \shadecell{0}{15.00} & \shadecell{0}{ 5.00} \\
		Pebbles    & \shadecell{0}{ 0.00} & \shadecell{0}{19.40} & \shadecell{1}{28.36} & \shadecell{0}{67.16} \\
		Rock       & \shadecell{0}{ 0.63} & \shadecell{0}{ 0.63} & \shadecell{0}{ 1.89} & \shadecell{1}{96.86} \\
		\bottomrule
	\end{tabular*}
	\caption{Confusion matrix and accuracies for training with \ac{IMU}. Total
		accuracy during training was \SI{84.02}{\%} and \SI{79.35}{\%} on the
		test set.}
	\label{tab:nn-res-imu}
\end{table}

Using the \ac{FTS} data as input to our \ac{NN} (\autoref{tab:nn-res-fts}), this
misclassification of `pebbles' as `rock' is diminished to \SI{10.45}{\percent}.
The overall test accuracy improves by almost 20 percentage points to
\SI{96.17}{\percent}. \autoref{tab:nn-res-combined} finally shows the best
performing \ac{NN}, trained and tested on both \ac{IMU} and \ac{FTS} data. While
the overall test accuracy only increases by ca.~0.1 percentage point, we can see
that the `pebbles' accuracy increases from \SI{88.06}{\percent} to
\SI{95.52}{\percent}. The negative effect of including the \ac{IMU} data back
in, is that the accuracy of the `compressed' class decreases from
\SI{98.75}{\percent} to \SI{95}{\percent}.

\begin{table}[htb]
	\begin{tabular*}{\linewidth}{@{\extracolsep\fill}lrrrr}%
		\toprule
		&\multicolumn{4}{c}{Predicted}\\\cmidrule{2-5}
		Actual & Loose & Compressed & Pebbles & Rock\\
		\midrule
		Loose      & \shadecell{1}{100.00} & \shadecell{0}{ 0.00} & \shadecell{0}{ 0.00} & \shadecell{0}{  0.00} \\
		Compressed & \shadecell{0}{  0.00} & \shadecell{1}{98.75} & \shadecell{0}{ 1.25} & \shadecell{0}{  0.00} \\
		Pebbles    & \shadecell{0}{  0.00} & \shadecell{0}{ 1.49} & \shadecell{1}{88.06} & \shadecell{0}{ 10.45} \\
		Rock       & \shadecell{0}{  0.00} & \shadecell{0}{ 0.63} & \shadecell{0}{ 1.89} & \shadecell{1}{ 97.48} \\
		\bottomrule
	\end{tabular*}
	\caption{Confusion matrix and accuracies for training with \ac{FTS}. Total
		accuracy during training was \SI{98.92}{\%} and \SI{96.17}{\%} on the
		test set.}
	\label{tab:nn-res-fts}
\end{table}

\begin{table}[htb]
	\begin{tabular*}{\linewidth}{@{\extracolsep\fill}lrrrr}
		\toprule
		&\multicolumn{4}{c}{Predicted}\\\cmidrule{2-5}
		Actual     & Loose                 & Compressed           & Pebbles                & Rock\\\midrule
		Loose      & \shadecell{1}{100.00} & \shadecell{0}{ 0.00} &  \shadecell{0}{  0.00} & \shadecell{0}{  0.00} \\
		Compressed & \shadecell{0}{  0.00} & \shadecell{1}{95.00} &  \shadecell{0}{  1.25} & \shadecell{0}{  3.75} \\
		Pebbles    & \shadecell{0}{  0.00} & \shadecell{0}{ 0.00} &  \shadecell{1}{ 95.52} & \shadecell{0}{  4.48} \\
		Rock       & \shadecell{0}{  0.63} & \shadecell{0}{ 0.63} &  \shadecell{0}{  1.26} & \shadecell{1}{ 97.48} \\
		\bottomrule
	\end{tabular*}
	\caption{Confusion matrix and accuracies for training with \ac{FTS} and
		\ac{IMU} data combined. Total accuracy during training was
		\SI{99.31}{\%} and \SI{96.76}{\%} on the test set.}
	\label{tab:nn-res-combined}
\end{table}

\begin{figure}[htb]
	\centering
	\begin{subfigure}[h]{0.32\linewidth}
		\includegraphics[width=\linewidth]{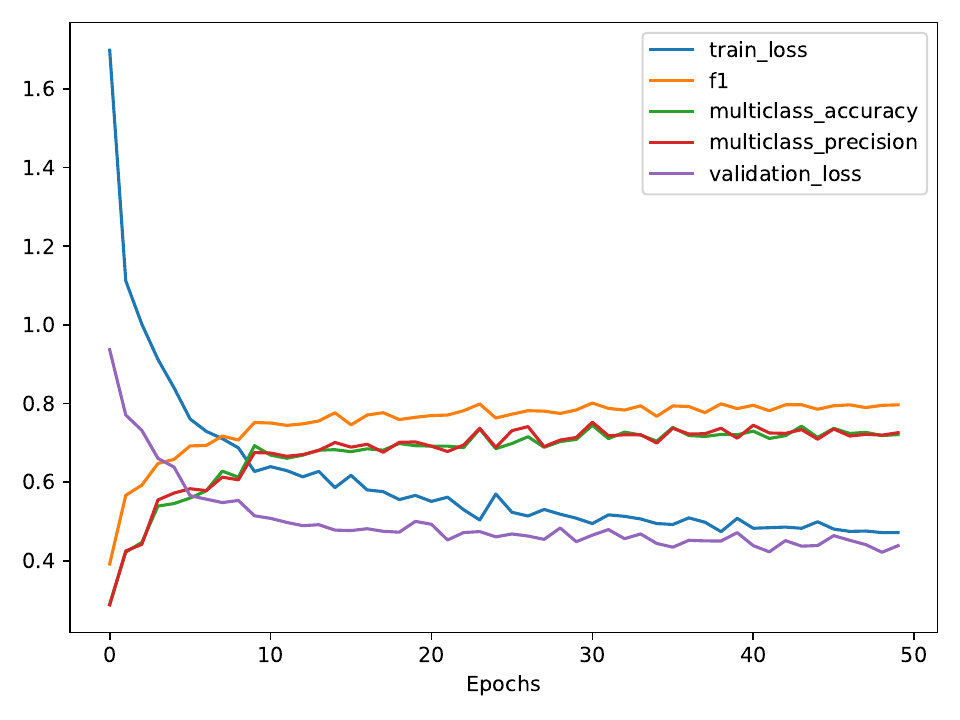}
		\caption{IMU}
		\label{fig:nn-imu}
	\end{subfigure}
	\begin{subfigure}[h]{0.32\linewidth}
		\includegraphics[width=\linewidth]{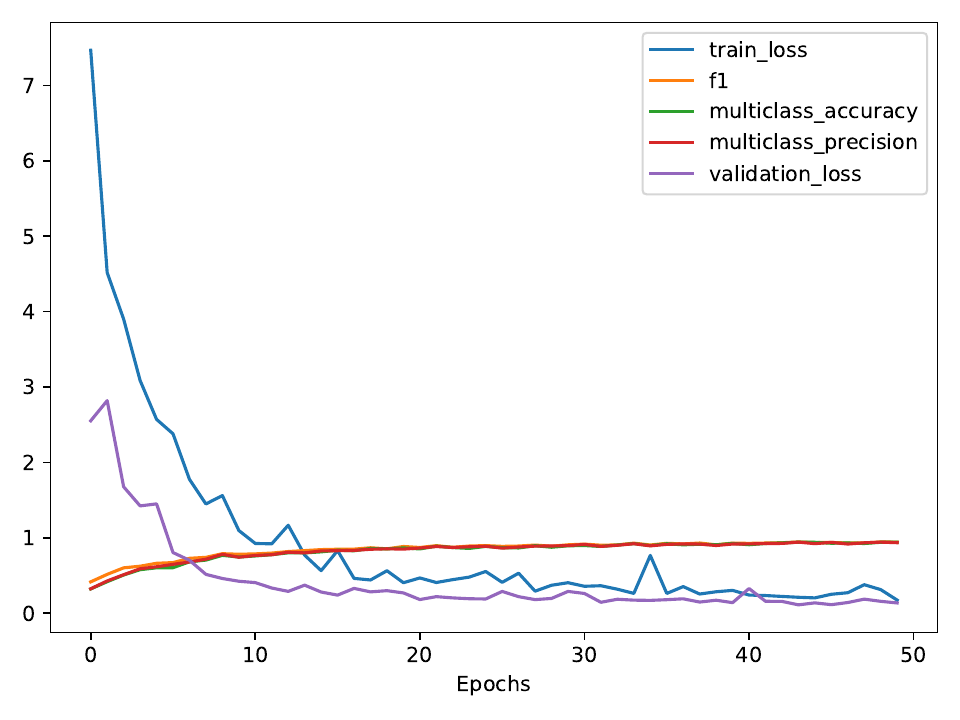}
		\caption{FTS}
		\label{fig:nn-fts}
	\end{subfigure}
	\begin{subfigure}[h]{0.32\linewidth}
		\includegraphics[width=\linewidth]{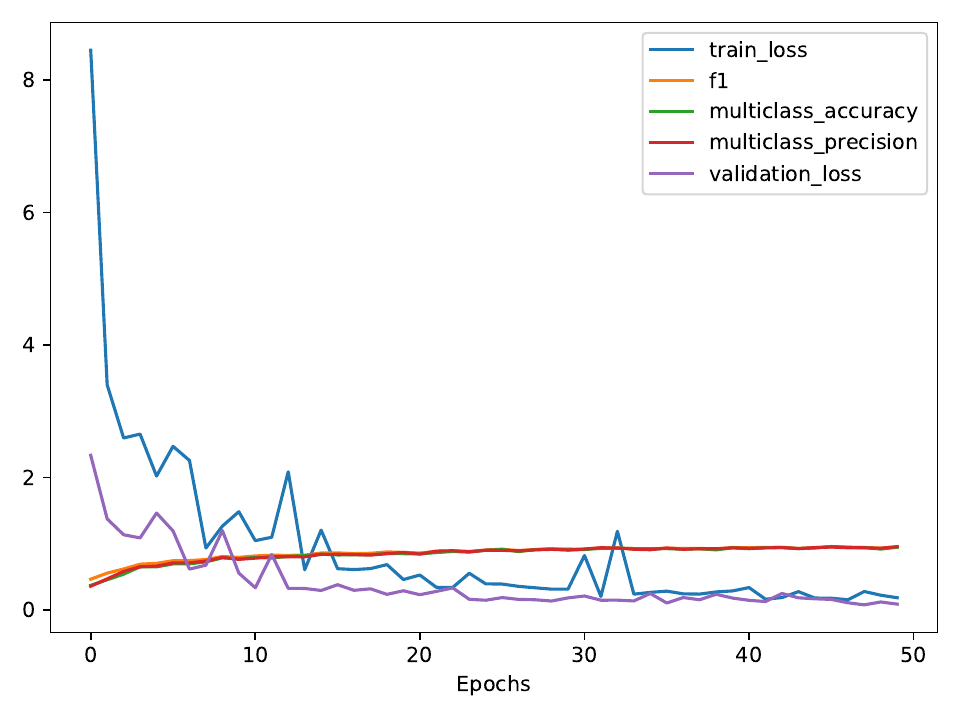}
		\caption{IMU and FTS}
		\label{fig:nn-all}
	\end{subfigure}
	\caption{Training curves for the \ac{NN} with (\subref{fig:nn-imu}) \ac{IMU}
		data, (\subref{fig:nn-fts}) \ac{FTS} data, and (\subref{fig:nn-all})
		both data sources combined.}
	\label{fig:nn-curves}
\end{figure}

\subsection{Classification summary}

From these results, we can conclude that the \ac{FTS} data provided superior
performance in the neural network classification task, especially for the
overlapping class `pebbles', for which also the visualization in the earlier
section struggled. This suggests that while \ac{IMU} data appears more effective
in lower-dimensional feature visualizations, the \ac{FTS} data outperforms it in
high-dimensional neural network applications.

One possible explanation is that, since \acp{NN} are an inherently data driven
approach, the \ac{FTS}-based approach exploits its ca.\ six times larger data
pool (remember that it is the same number of samples, but each sample now
contains statistics of six \ac{FTS} compared to just one \ac{IMU}). At the same
time, the individual \ac{FTS} is mounted at a leg and is connected to the rover
chassis only by proxy of passive, movable bogies. As such, it has a less
holistic view of the terrain compared to the chassis-mounted \ac{IMU} near the
center of mass and rotation. Considering the vertical force as an example, the
rover mass is distributed over all six legs at any given time and each leg
experiences a distinctly different vertical load depending on the current
orientation and rover configuration, i.e., bogie angles.

Training and inference times were measured for both \ac{SVM} and \ac{NN} on a
laptop with an 11th Gen Intel Core i7-11800H and a GeForce RTX 3050 Ti Mobile.
\autoref{tab:svm-fts-time} shows these timing metrics and the learning curves of
the \ac{NN} can be seen in  \autoref{fig:nn-curves}. We notice that the training
of the \ac{NN} with one set of hyperparameters takes just over \SI{30}{\sec} on
the GPU while the \ac{SVM} needs ca.\ \SI{10}{\sec} less to perform the entire
grid search for fitting hyper-parameters. By contrast, fitting the \ac{SVM} with
one set of parameters takes only \SI{0.13}{\sec}.

The surprising number is the inference time for the whole set. To achieve more
comparable results, we ran the \ac{NN} inference on the CPU, too. Notice that
the \ac{SVM} needs \SI{0.0076}{\sec}, more than five times as much time as the
\ac{NN}.

\begin{table}[htb]
	\centering
	\begin{tabular*}{\linewidth}{@{\extracolsep\fill}lccc}\toprule
		& Training grid search & Training (best params) & Inference \\\midrule
		SVM & 24.25522             & 0.132472               & 0.00762   \\
		NN  & N/A                  & 33.751                 & 0.0014288 \\\bottomrule
	\end{tabular*}
	\caption{Average training and inference times [s] for the \ac{SVM} and \ac{NN}
		on a 11th Gen Intel Core i7-11800H and a GeForce RTX 3050 Ti Mobile when
		using all available data and an a \SI{25}{\percent} test split. The
		averages are over five repeats and inference time is inference for the
		whole test set, i.e.\ 339 samples, on the CPU.}
	\label{tab:svm-fts-time}
\end{table}

This could be due to the properties of our classification dataset. Since we only
considered traverse sections that are only one or the other class for a
relatively long time, this dataset is small but high-dimensional.

Both \ac{SVM} and \ac{NN} are fast enough for us to consider them for use on the
rover, because we do not foresee running a classification with more than
\SI{1}{Hz}, especially considering that some of the statistics are computed over
\SI{1}{\sec}-long windows and, more importantly, our rovers generally move
slower than \SI{1}{\meter\per\sec}.

\FloatBarrier
\section{Drawbar Pull Estimation}

In their works, Wong et al.\ \cite{Wong2001} as well as Ishigami et al.\
\cite{Ishigami2010} stipulate the relationship between drawbar pull, vertical
and lateral forces, wheel radius, width, normal stress beneath the wheel, shear
stresses along the longitudinal and lateral wheel directions, the entry and exit
angle, reaction resistances, and the angle to the principal ground contact
point. Note that wheel deformation and dynamics are not mentioned yet, but we
can already see the interest in finding a direct way to measure the drawbar pull
instead of inferring it from these, mostly, external factors. Instead, with a
known drawbar pull, we could use statistical approaches to inform about some of
those unknown ground parameters \cite{Ishigami2010}.

The influence of both vibrations and noise renders it difficult to or in parts
even inhibits us from interpreting force and torque signal over the entire
traverses. The estimated drawbar pull could sometimes vary between tens of
Newtons within seconds even for flat terrains and straight traverses. Instead,
we propose to filter the measured data using the known rover geometry.

\begin{figure}[htb]
	\centering
	\includegraphics[width=0.3\linewidth]{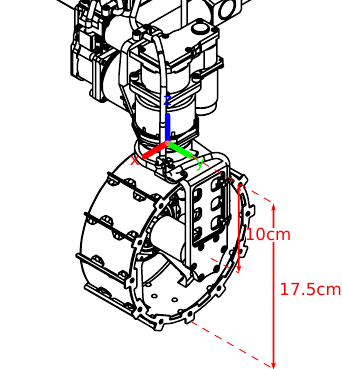}
	\caption{The vertical distance between the \ac{FTS} and the wheel axis is
		\SI{10}{\cm}. With a wheel diameter of \SI{15}{\cm}, this results in a
		maximum distance of \SI{17.5}{\cm} from \ac{FTS} to the ground contact
		point. On the left side of the rover, the \ac{FTS}'s $x$ axis points
		forward, $y$ to the left (away from the rover chassis), and $z$ upward.
		On the rover's right-hand side, the \acp{FTS} are rotated
		\SI{180}{\degree} around $z$, resulting in $x$ pointing backwards.}
	\label{fig:fts-phys-distances}
\end{figure}

We know that the normal distance from the \ac{FTS} to the ground contact point
is \SI{0.175}{\m} and that the distance between the sensor and the wheel axle is
\SI{0.1}{\m} (see \autoref{fig:fts-phys-distances}). We can view this subsystem
as a lever of length $L$ of \SIrange{0.1}{0.175}{\meter} (\autoref{eq:dbp}),
depending on where exactly ground contact is made, where the torque at the
\ac{FTS}, $\tau_y$, depends on the horizontal force and the vertical distance to
the contact point.

\begin{equation}
	\label{eq:dbp}
	\tau_y = F_x \cdot L
\end{equation}

The principal assumptions are a rigid linkage between the sensor and the wheel,
a rigid wheel, steering straight on a flat terrain, no sinkage, and no grousers.
The terrains that we traversed did not show large sinkage, but to account for
the real terrain and system, in this first attempt, we propose to use a generous
threshold while filtering the \ac{FTS} signals to only those values where the
measured force and torque yield a distance $L$ within said thresholds of
\SIrange{10}{17.5}{\cm} to discard noise and find more easily interpretable
intervals of the traverses.

\begin{figure}[htb]
	\includegraphics[width=\linewidth]{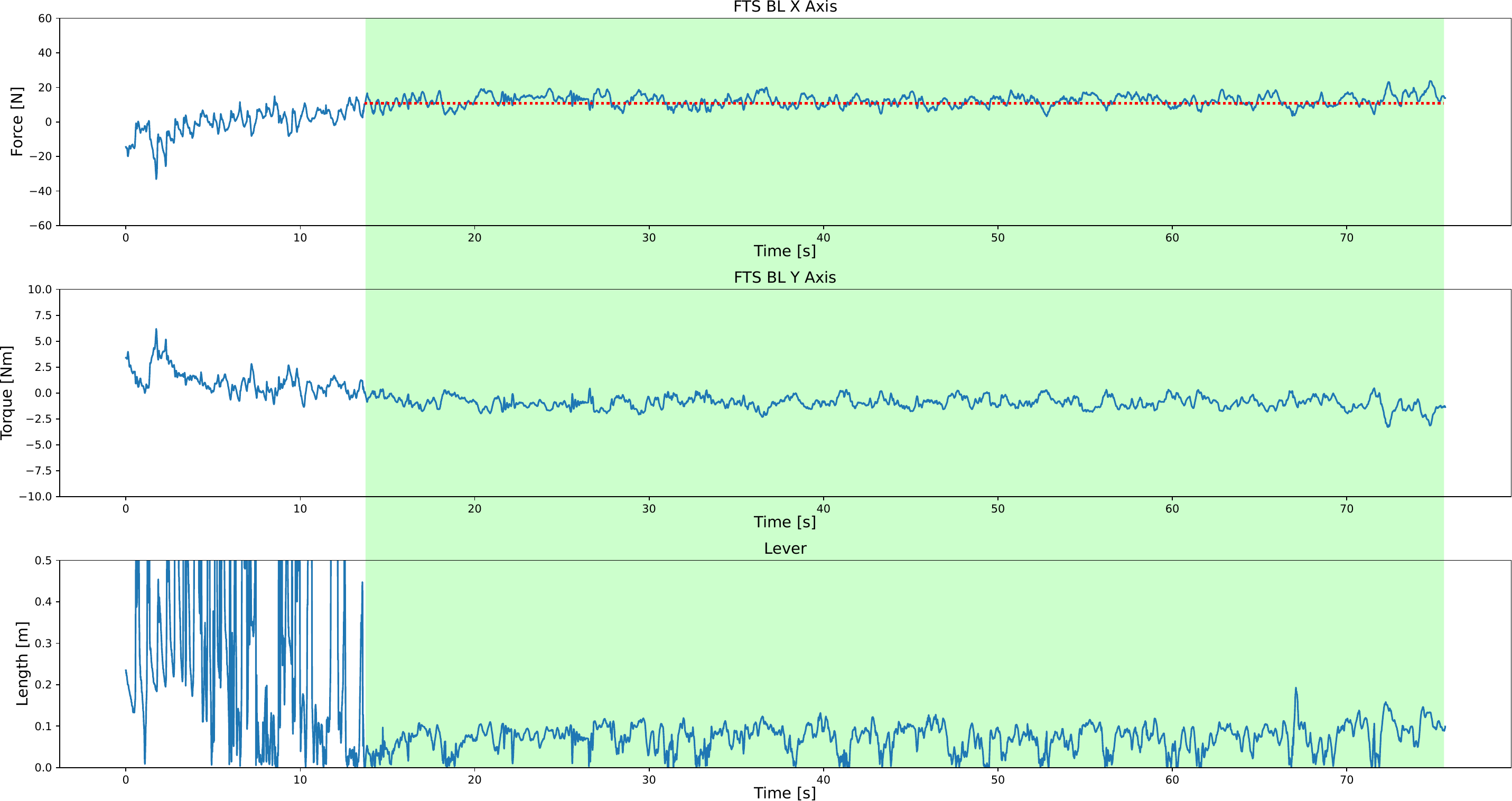}
	\caption{Rear left FTS data for a traverse over loose soil, extracted from
		traverse `2023-07-21 17-34-18'. The manually highlighted area
		is where the lever length appears stable.}
	\label{fig:fts-loose-long}
\end{figure}

\begin{figure}[htb]
	\includegraphics[width=\linewidth]{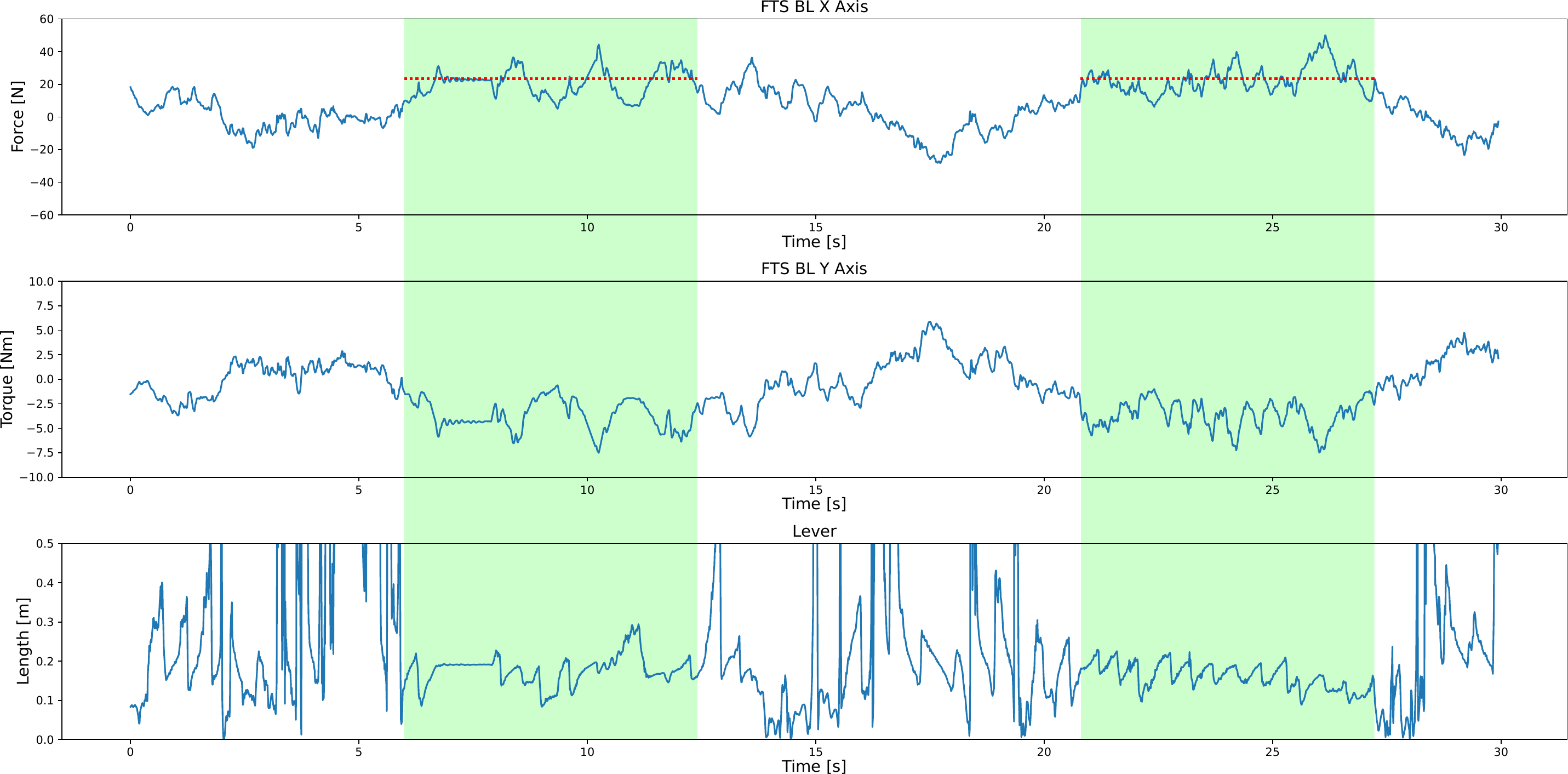}
	\caption{Rear left FTS data of a traverse over compressed sand, extracted
		from traverse `2023-07-20 18-12-05'. The manually highlighted areas are
		where the lever length appears stable.}
	\label{fig:fts-compressed-long}
\end{figure}

\begin{figure}[htb]
	\includegraphics[width=\linewidth]{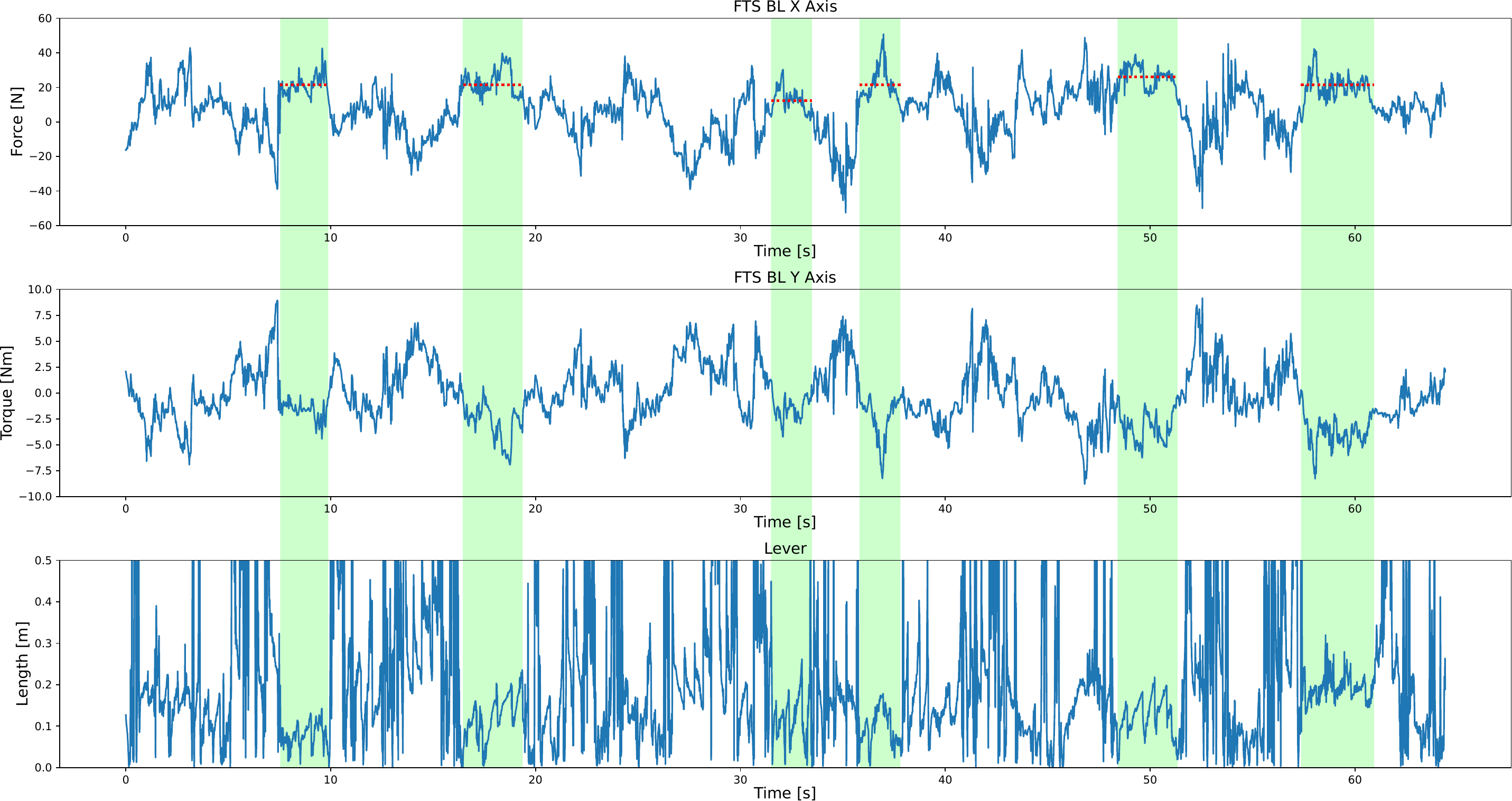}
	\caption{Rear left FTS data for a traverse over a rocky outcrop,
		extracted from traverse `2023-07-21 12-58-11'. The manually highlighted
		areas are where the lever length appears stable.}
	\label{fig:fts-rock-long}
\end{figure}

\autoref{fig:fts-loose-long}, \autoref{fig:fts-compressed-long}, and
\autoref{fig:fts-rock-long} show traverse data for three terrain types loose,
compacted, or rock, respectively. \autoref{fig:terrains} shows these three
terrains from the rover camera perspective. We have highlighted the time
intervals where the computed lever length could be considered stable and close
to the proposed interval of \SIrange{10}{17.5}{\cm} with considerable margin.
Because we do not have an exact ground truth, we cannot tune additional filters.
What these graphs convey already, however, is that a restriction of the
measurements down to stable lever length values within an interval with
geometrical meaning can be a first sensible approximation. It can be seen that
the soft soil example within the highlighted area has a lower variance than
outside of it. The force within the interval is close to \SI{10}{N}. In the
harder, compressed terrain, we did not find stable intervals as long as in the
soft soil example. Where we did highlight them, the force in $x$ direction was
closer to \SI{20}{N}. The last example, traverse over rock, has a higher
variation and we cannot identify intervals as long as for the other two example
traverses. For the manually identified intervals, the values are between
\SI{10}{N} and \SI{20}{N}. Recalling the data inspection for the classification,
this is visible in the \ac{IMU} and \ac{FTS}'s $z$ axis readings, too. The rover
experiences more vibrations and the grousers cannot dig into the ground, pushing
the wheels up at every contact instead.

\begin{figure}[htb]
	\centering
	\begin{subfigure}[h]{0.32\linewidth}
		\includegraphics[width=\linewidth]{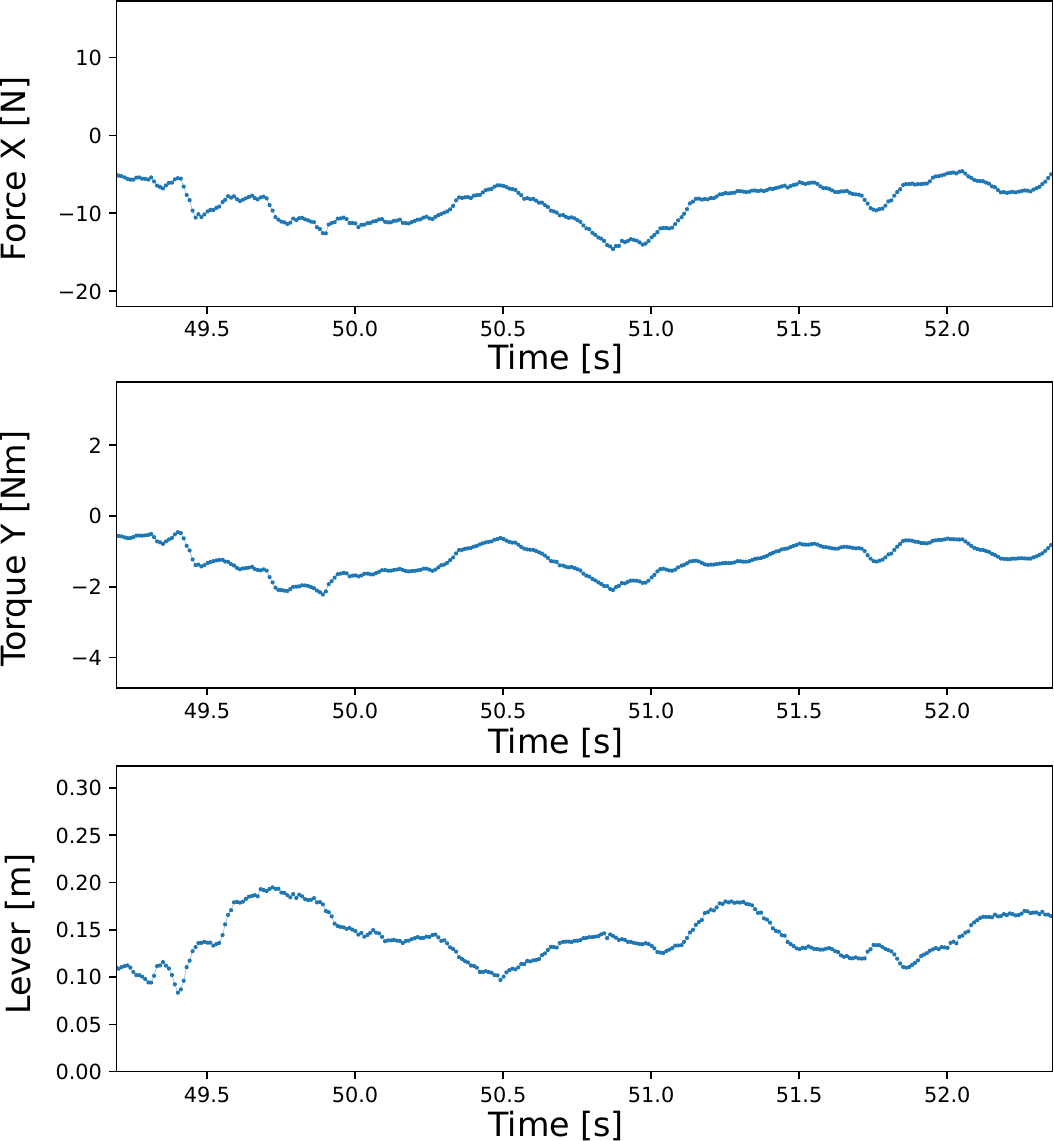}
		\caption{\raggedright Loose soil, extracted from traverse `2023-07-21 17-34-18'.}
		\label{fig:fts-loose-short}
	\end{subfigure}
	\begin{subfigure}[h]{0.32\linewidth}
		\includegraphics[width=\linewidth]{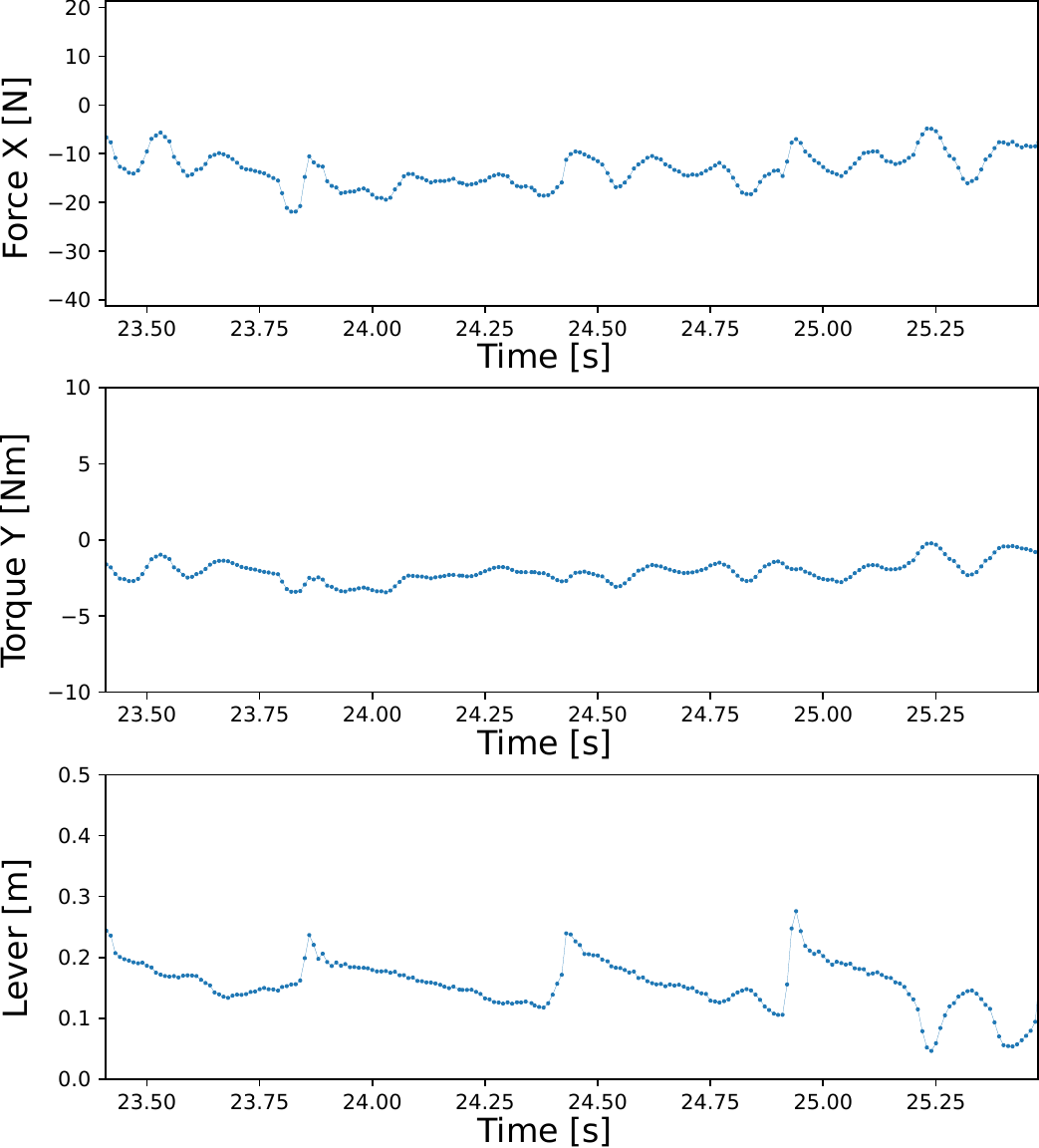}
		\caption{\raggedright Compressed sand, extracted from traverse `2023-07-20 18-12-05'.}
		\label{fig:fts-compressed-short}
	\end{subfigure}
	\begin{subfigure}[h]{0.32\linewidth}
		\includegraphics[width=\linewidth]{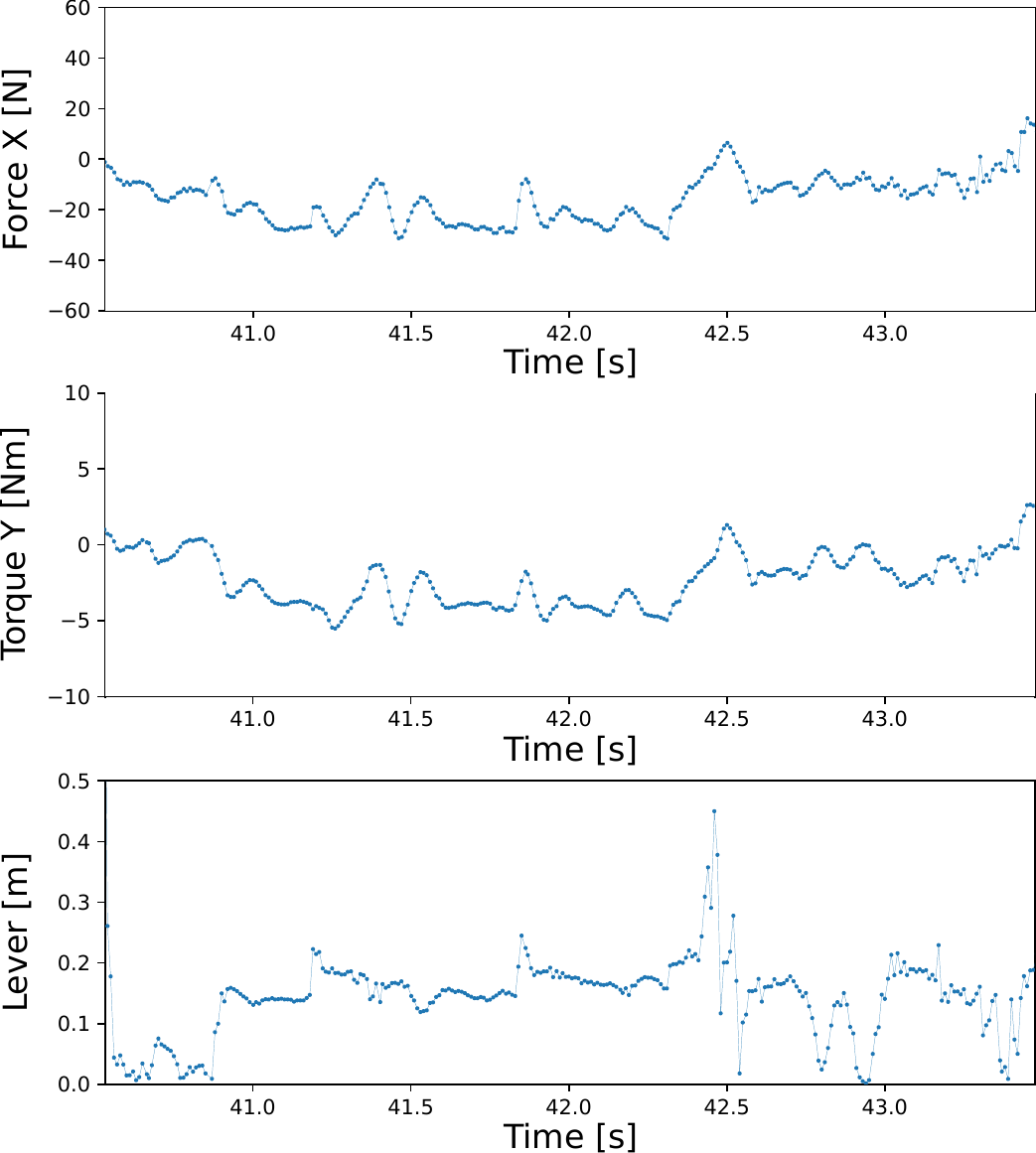}
		\caption{\raggedright Rocky outcrop, extracted from traverse `2023-07-21 12-58-11'.}
		\label{fig:fts-rock-short}
	\end{subfigure}
	\caption{FTS data for traverses over different terrain types, zoomed in on
		time slots where the computed lever is within ca.~\SIrange{10}{17.5}{\cm}.}
	\label{fig:fts-short}
\end{figure}

\autoref{fig:fts-short} shows the same terrains as figures
\ref{fig:fts-loose-long}, \ref{fig:fts-compressed-long}, and
\ref{fig:fts-rock-long}, but zoomed in on three short time intervals where the
computed lever length stays within the threshold above. It appears that the
forces in $X$ direction is \SIrange{5}{10}{N} for the loose soil sample, moves
around \SI{15}{N} $\pm$ \SI{5}{N} for the compressed sand sample, and is much
more variable for the rocky terrain.

\newcommand{\round}[1]{\num[round-mode=places, round-precision=2]{#1}}
\begin{table}[htb]
	\centering
	\begin{tabular*}{\linewidth}{@{\extracolsep\fill}cccccccc}\toprule
		& \multicolumn{7}{c}{Remaining [\%]}   \\\cmidrule{2-8}
		Tolerance [cm] & FL & FR & CL & CR & BL & BR & Total  \\
		\midrule
		5
		& \round{75.0525841627202}
		& \round{61.92724187161598}
		& \round{57.79915057533705}
		& \round{59.13159589158716}
		& \round{74.65048569842251}
		& \round{55.04919275395991}
		& \round{63.93503731088278} \\
		2
		& \round{61.31911983352651}
		& \round{46.656834487155313}
		& \round{41.557645196096477}
		& \round{43.61572654279672}
		& \round{60.97984475010214}
		& \round{39.91035047569021}
		& \round{49.00658122169384} \\
		1
		& \round{54.63545014362903}
		& \round{40.379429344491435}
		& \round{34.74320467752242}
		& \round{37.60160008560129}
		& \round{53.99704136255954}
		& \round{33.99700619525279}
		& \round{42.55894939897029} \\

		\bottomrule
	\end{tabular*}

	\caption{Filtering \ac{FTS} data to tolerances according to the first column
		around \SIrange{10}{17.5}{\cm} (e.g., \SIrange{9}{18.5}{\cm}),
		the following columns show how many data points [\SI{}{\percent}]
		remain for the respective \ac{FTS} and overall.}

	\label{tab:valid_fts_percentages}
\end{table}

\begin{figure}[htb]
	\centering
	\includegraphics[width=.75\linewidth]{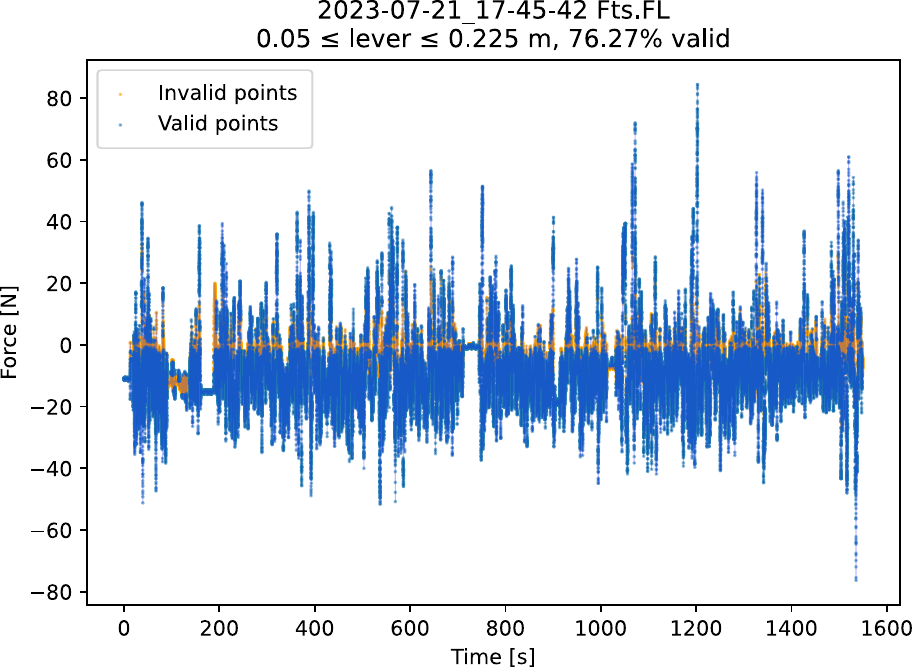}
	\caption{This scatter plots show the result of filtering by computed lever
		arm length on force and torque ($F_x$, $\tau_y$) over one traverse.
		The blue points are left after filtering while orange are the points that
		are removed. We can see that values close to 0 are removed because $F_x$
		is in the denominator and we remove some instances where the rover is
		standing still (left), but not all (center).}
	\label{fig:filtered_fts}
\end{figure}

\FloatBarrier
\section{Conclusions}\label{sec5}

We have presented the different use cases for \acfp{FTS} on mobile robot bases
in the context of planetary exploration. A rover testbed equipped with six
\acp{FTS} as well as a more traditional \acf{IMU} was used in a field test to
evaluate the applicability of the sensors for these scenarios.

Mounting the \acp{FTS} above the wheel potentially simplifies wiring and
maintenance. The main argument for mounting the \acp{FTS} above the wheel is the
additional characterization of alternative locomotion modes such as wheel
walking, where the use of the deployment/walking joints is of interest. However,
to gain more insights into the interaction between the rover and the terrain
while driving, we recommend installing them in the drive hub and correlating
them with the motor current. This allows for a more direct measurement of the
effort at the wheel.

Working with the sensor data captured during long traverses over varying terrain
types, slopes, and at varying speeds, we have experienced that drawbar pull
cannot immediately be read from the \ac{FTS} in a meaningful way. We have,
however, identified a candidate approach for identifying time intervals in which
the drawbar pull could be derived from the force-torque sensors. Additionally,
the proposed approach has a geometrical meaning such that its parameters can be
inferred from the rover's legs. Note though that the confirmation and tuning of
additional filtering could not be done with field test data alone. For this, we
would need to design experiments with external ground-truth measurements of the
drawbar pull.

While invaluable for use in manipulators, the case for \acp{FTS} in the wheeled
mobile base is harder to argue. In more controlled environments such as single
wheel testbeds, they prove useful, but as the vibrations are superimposed on the
signals, the \acp{FTS} become harder to interpret as drawbar pull estimations.
Having gone through these trials ourselves, we recommend foregoing comparatively
costly \acp{FTS} and rather use cheaper and conventional sensors in their stead.
Especially if the rover in question only uses more traditional driving modes.
For slip detection, we will choose current sensors in the future or at least
mount the \acp{FTS} in the wheel hub itself instead of above the wheel. For
terrain classification, the \acp{FTS} proved better than the \ac{IMU}, but we
believe that this can be mitigated by using an additional \ac{IMU} as discussed
earlier. The main remaining argument for the existing \ac{FTS} configuration we
see, is that wheel walking performance could theoretically be evaluated based on
only one instead of two \acp{FTS} per leg.

\section{Future work}

We have explored the most prominent use cases of \acp{FTS} for rover navigation
above, but we see additional tasks that should be explored in the future to
fully exploit the sensor mounting location for rover navigation.

As demonstrated in the section about drawbar pull, we have identified a
candidate for selecting valid time intervals. This needs to be confirmed and
supported by additional filtering by tests with ground-truth data. The tests
should include an external measurement of drawbar pull, different, defined
slopes with known ground parameters, a wheel with and without grousers, and a
complete formulation of rover kinematics and dynamics. The latter is especially
interesting for planetary rovers which typically feature articulated bogies and
steering joints.

The orientation of the rover, roll and pitch in particular, could be estimated
using the force torque data. This estimate can then be compared to \ac{IMU} data
and current kinematics transformations to get a more complete image of the rover
configuration and inform the confidence estimates of individual sensors for any
sensor fusion.

For configurations such as the ExoMars rover \cite{Patel2010}, the exerted
torques at the deployment actuators are of interest, specifically during wheel
walking analysis \cite{WheelWalkingAstra,TimWheelWalking}. Again, this can be
measured directly at the joint, but if we want to measure both at this joint and
at the drive, we would end up with two sensors. In the MaRTA rover, which is a
half-scale model of ExoMars, however, the decision was taken to install only one
\ac{FTS} per leg which could still deliver approximations for both
deployment/walking and driving, yielding indicative results for both.

To validate the applicability of the \ac{FTS} in \ac{MaRTA}'s configuration for
determining the drawbar pull, a single-wheel testbed should be equipped with an
\acs{FTS} about the wheel and the real drawbar pull measured for different
terrains. An additional difficulty in determining the drawbar pull contribution
of each wheel consists in the rover configuration and pose. As can be seen in
\autoref{fig:kinematics-abstract}, each wheel can have its own pose instead of
always being aligned with the rover chassis. This is in addition to the chassis
orientation and steering angles. This could also lead to the integration of
traction control.

\backmatter
\section*{Statements and declarations}
\subsection*{Data availability}

This paper is based on the dataset from our Bardenas field test, BASEPROD. This
dataset is documented in \cite{Baseprod} and available for download at
\cite{BaseprodRepo}. The source code for the plots and terrain classification
are available at \url{https://github.com/spaceuma/fts-assessment}.

\subsection*{Funding}

\hbadness=1331
This work was supported by the European Space Agency under activity no.\
4000140043/22/NL/GLC/ces.

\subsection*{Conflict of interest}

The authors have no relevant financial or non-financial interests to disclose.

\subsection*{Author contributions}

The study uses published data from a field test to which all authors
contributed. All authors contributed to the study's conception and design. The
material was prepared by Levin Gerdes and signals were analyzed by Levin Gerdes
conceptualized and prepared by Levin Gerdes. The first draft of the manuscript
was written by Levin Gerdes and all authors commented on previous versions of
the manuscript. All authors read and approved the final manuscript.

\bibliography{references.bib}

\end{document}